\documentclass[11pt]{article}
\usepackage{graphicx}
\usepackage{fullpage}
\usepackage{amsmath,amssymb,amsthm}
\usepackage{enumerate}
\usepackage{mathrsfs}
\usepackage{bbm}
\usepackage{verbatim}
\usepackage{marginnote} 
\usepackage[T1]{fontenc}
\usepackage{kpfonts}
\usepackage{microtype}
\usepackage{enumitem}

\numberwithin{equation}{section}
\newtheorem{theorem}{Theorem}[section]
\newtheorem{lemma}{Lemma}[section]

\newtheorem{proposition}{Proposition}[section]

\newtheorem{remark}{Remark}[section]

\usepackage[usenames]{color}
\definecolor{plum}  {rgb}{.4,0,.4}
\definecolor{BrickRed} {rgb}{0.6,0,0}
\usepackage{commath}
\usepackage[normalem]{ulem}

\usepackage[plainpages=false,pdfpagelabels,colorlinks=true,linkcolor=BrickRed,citecolor=plum]{hyperref}
\usepackage{cleveref}

\usepackage{amsopn}

\def\ddefloop#1{\ifx\ddefloop#1\else\ddef{#1}\expandafter\ddefloop\fi}
\def\ddef#1{\expandafter\def\csname c#1\endcsname{\ensuremath{\mathcal{#1}}}}
\ddefloop ABCDEFGHIJKLMNOPQRSTUVWXYZ\ddefloop
\def\ddef#1{\expandafter\def\csname s#1\endcsname{\ensuremath{\mathsf{#1}}}}
\ddefloop ABCDEFGHIJKLMNOPQRSTUVWXYZ\ddefloop
\def\Pr{\mathbf{P}}
\def\E{\mathbf{E}}

\def\Reals{\mathbb{R}}
\def\argmin{\operatornamewithlimits{arg\,min}}
\def\deq{:=}
\def\wh#1{\widehat{#1}}
\def\eps{\varepsilon}

\def\diam{\mathsf{diam}}

\def\comp{\mathfrak{C}}

\def\risk{\cR}
\def\d{\mathrm{d}}
\def\conv{\mathsf{conv}}\include{learncode.tex}

\def\gen{\mathsf{gen}}

\ifpdf
\hypersetup{
  pdftitle={Learning finite-dimensional coding schemes with nonlinear reconstruction maps},
  pdfauthor={J. Lee and M. Raginsky}
}
\fi

\title{Learning finite-dimensional coding schemes with nonlinear reconstruction maps}

\author{Jaeho Lee\thanks{University of Illinois; {\tt jlee620@illinois.edu}.} \and
Maxim Raginsky\thanks{University of Illinois; {\tt maxim@illinois.edu}.}}

\date{}

\begin{document}
\maketitle

\begin{abstract}
This paper generalizes the Maurer--Pontil framework of finite-dimensional lossy coding schemes to the setting where a high-dimensional random vector is mapped to an element of a compact set of latent representations in a lower-dimensional Euclidean space, and the reconstruction map belongs to a given class of nonlinear maps. Under this setup, which encompasses a broad class of unsupervised representation learning problems, we establish a connection to approximate generative modeling under structural constraints using the tools from the theory of optimal transportation. Next, we consider problem of learning a coding scheme on the basis of a finite collection of training samples and present generalization bounds that hold with high probability. We then illustrate the general theory in the setting where the reconstruction maps are implemented by deep neural nets.
\end{abstract}

\section{Introduction}\label{sec:intro}
The problem of \textit{lossy compression} is about constructing succinct representations of high-dimensional random vectors that retain the features of the data that are relevant for some subsequent task, such as reconstruction subject to a fidelity criterion or statistical inference. When the compressed representation is digital, with constraints imposed by the limitations on the speed of digital transmission or on the available storage space, the corresponding problem of lossy compression falls within the purview of rate-distortion theory \cite{berger_71} and the theory of vector quantization \cite{gershogray_12}. On the other hand, given recent advances in machine learning using deep neural nets \cite{goodfellowetal_16}, it is of interest to consider `analog' schemes for lossy compression that map the original high-dimensional data to a \textit{continuous} latent representation of lower dimensionality \cite{bengioetal_13}, and where the reconstruction operations that send the latent representation back to the original high-dimensional space are implemented by nonlinear maps with a given structure. Moreover, even if one can show the existence of an \textit{optimal} coding scheme matched to a given data-generating distribution, this distribution is often unknown, and one has to resort to \textit{empirical design} (or learning) of coding schemes on the basis of training samples. This approach encompasses both classical problems like clustering and vector quantization \cite{gershogray_12,linder_02} or principal component analysis \cite{jolliffe_02}, and modern \textit{unsupervised representation learning} techniques, such as autoencoders \cite{ollivier_14}. In all of these scenarios, it is of interest to obtain theoretical bounds on the optimality gap (or generalization error) of the learned coding scheme.


Recently, Maurer and Pontil \cite{maurerpontil_10} studied the problem of learning finite-dimensional coding schemes with compact low-dimensional representation spaces and linear reconstruction maps and used empirical process techniques to derive the bounds on the generalization error. Follow-up work by Vainsencher et al.~\cite{vainsencheretal_11} extended the results of \cite{maurerpontil_10} to the setting of dictionary learning. In this paper, we consider the problem of learning finite-dimensional coding schemes with low-dimensional compact representation spaces and nonlinear reconstruction maps, such as deep neural nets. Moreover, the utility of finite-dimensional coding schemes is not limited to compression --- one can also view them as approximate \textit{generative models} for a given signal class subject to suitable structural constraints. For example, it was shown by Pollard \cite{pollard_82} that an optimal $k$-point vector quantizer for a $d$-dimensional random vector $Z$ can be turned into a generative model that best approximates the probability law of $Z$ by a discrete probability measure supported by $k$ points in $\Reals^d$ (if $q$ is the map that implements the quantizer, then the probability law of $q(Z)$ gives the best approximation of the law of $Z$ in the Euclidean $2$-Wasserstein metric --- see \cref{sssec:transport} for a detailed discussion). One of the contributions of this paper is to show that this generative viewpoint is valid for a much wider class of lossy compression schemes with nonlinear reconstruction maps, e.g. deep neural networks.

The remainder of the paper is organized as follows. In \cref{sec:framework}, we present a comprehensive theoretical framework of finite-dimensional coding schemes and discuss its close relation to unsupervised learning of latent representations. We also discuss its connection to optimal transport theory \cite{villani_03} and rate-distortion theory \cite{berger_71}. In particular, the optimal transport viewpoint, detailed in \cref{sssec:transport}, provides the foundation for viewing finite-dimensional coding schemes as approximate generative models for high-dimensional data subject to structural constraints. In \cref{sec:learn}, we formulate the problem of empirical design or learning of a coding scheme and provide two bounds on the generalization error, one based on the theory of optimal transport and another based on exploiting the geometric complexity of the class of reconstruction maps. In \cref{sec:stack}, we exemplify the use of the latter generalization bound in the context of finite-dimensional coding schemes with reconstruction maps implemented by deep neural nets composed of fully-connected layers or convolutional layers. All proofs are relegated to \cref{sec:proof}.

\paragraph{Notation}
For a vector, $\|\cdot\|$ denotes the $\ell_2$ norm unless specified otherwise. For a matrix $A$, $\|A\|$ denotes the spectral norm: $\|A\| \deq \sup \{\|Au\| : \|u\|=1\}$. For $p \geq 1$, the norm $\|\cdot\|_p$ for matrices denotes the entrywise $\ell_p$ norm, i.e., for an $m \times n$ matrix $A$, $\|A\|_p \deq \left(\sum_{j=1}^n \sum_{i=1}^m |a_{ij}|^p\right)^{1/p}$. For a set $\cU$ of vectors, $\|\cU\|_{\infty}$ denotes the maximum $\ell_2$ norm of the elements in $\cU$, i.e. $\sup_{u \in \cU}\|u\|$. We will use the standard $\cO(\cdot)$ notation, and will use  $\tilde{\cO}(\cdot)$ to hide logarithmic factors. All logarithms are taken to base $e$.

\section{The framework of $k$-dimensional coding schemes}\label{sec:framework}

We consider a class of coding schemes for a random vector $Z$ taking values in a subset $\cZ$ of $\Reals^d$. A \textit{$k$-dimensional coding scheme} (with $k \le d$) consists of a compact set $\cH \subset \Reals^k$ (which will be referred to interchangeably as the \textit{codebook}, the \textit{latent space}, or the \textit{representation space}) and a measurable map $f : \cH \to \Reals^d$ (the \textit{reconstruction map}) which is an element of a given class $\cF$ of admissible reconstruction maps. The \textit{reconstruction error} of $h \in \cH$ for $Z$ is defined as
\begin{align}
	e_f(Z,h) \deq \| Z - f(h) \|^2,
\end{align}
and we consider the \textit{minimal reconstruction error}
\begin{align}\label{eq:min_error}
	e_f(Z) \deq \min_{h \in \cH} e_f(Z,h) = \min_{h \in \cH} \| Z - f(h) \|^2.
\end{align}
We assume enough regularity for the elements of $\cF$ to guarantee the existence of the minimum in \eqref{eq:min_error} --- since $\cH$ is compact, it suffices to ensure that all functionals of the form $h \mapsto e_f(z,h)$ ($z \in \cZ$, $f \in \cF$) are lower semicontinuous. Let $P$ denote the probability law of $Z$. Then the \textit{expected reconstruction error} of $f \in \cF$ is given by
\begin{align}
	\risk(P,f) = \E_P [e_f(Z)] = \E_P \left[\min_{h \in \cH}\|Z - f(h)\|^2\right]. \label{eq:representation_risk}
\end{align}
Given the class $\cF$, an \textit{optimal coding scheme} for $P$ is any element $f\in\cF$ that attains the \textit{minimum reconstruction error} $\risk(P,\cF) \deq \inf_{f \in \cF} \risk(P,f)$. In this sense, learning coding schemes can be understood as an unsupervised statistical learning problem with \textit{induced hypothesis space} consisting of the minimal error functions $e_f$ for all $f \in \cF$.

\subsection{Relationship to representation learning frameworks}

This framework is closely related to the notion of $k$-dimensional coding schemes introduced by Maurer and Pontil \cite{maurerpontil_10}. In that work, $Z$ is a random element of the unit ball of a (possibly infinite-dimensional) Hilbert space ${\mathbb H}$, the codebook $\cH$ is a compact subset of $\Reals^k$, and $\cF$ is taken to consist of linear operators $f : \Reals^k \to {\mathbb H}$ obeying the constraint
$$
\sup_{f \in \cF} \sup_{h \in \cH} \| f(h) \|_{{\mathbb H}} < \infty.
$$
Here, we restrict ${\mathbb H}$ to be finite-dimensional, but allow nonlinear reconstruction maps.

This extension enables us to treat modern variants of unsupervised representation learning, such as autoencoders \cite{ollivier_14}, under the same framework as vector quantization or $k$-means clustering, principal component analysis (PCA), nonnegative matrix factorization, and sparse coding, by carefully selecting the latent space $\cH$ and the class of reconstruction maps $\cF$. We present three simple illustrative examples below:

\paragraph{Vector quantization} A $k$-point vector quantizer on $\Reals^d$ is specified by a codebook $\cC = \{u_1,\ldots,u_k\} \subset \Reals^d$ and the (nearest-neighbor) encoding map
\begin{align}
	Z \mapsto \argmin_{1 \le j \le k} \| Z - u_j \|^2,
\end{align}
with a fixed but arbitrary tie-breaking rule. The reconstruction error is given by $e_\cC(z) = \min_{1 \le j \le k} \|z - u_j\|^2$. As shown by Maurer and Pontil \cite{maurerpontil_10}, vector quantization is an instance of a linear $k$-dimensional coding scheme with $\cH = \{e_1,\ldots,e_k\}$ (the canonical orthonormal basis of $\Reals^k$) and the linear reconstruction map $f : \Reals^k \to \Reals^d$ defined by $f(e_j) \deq u_j$ ($1 \le j \le k$) and extended to all of $\Reals^k$ by linearity. Indeed, by the construction of $f$,
\begin{align}
	e_f(z) = \min_{h \in \cH} \| z - f(h) \|^2 = \min_{1 \le j \le k} \| z - f(e_j) \|^2 = \min_{1 \le j \le k} \| z - u_j \|^2 = e_\cC(z).
\end{align}

\paragraph{Principal Component Analysis} In principal component analysis (PCA), one aims to construct a projection operator which maps vectors in the observation space $\Reals^d$ to a $k$-dimensional linear subspace $\cK$. The objective is to find a projection operator $\Pi : \Reals^d \to \Reals^d$ with $k$-dimensional range to maximize the \textit{energy} of the projected vector $\E\|\Pi Z\|^2$. From the definition of projection and the fact that any projection can be decomposed as $\Pi = T T^*$ for some linear isometry $T : \Reals^d \to \Reals^k$ and its adjoint $T^* : \Reals^k \to \Reals^d$, we have
\begin{align*}
	\|\Pi Z\|^2 = \|Z\|^2 - \|Z - \Pi Z\|^2 = \|Z\|^2 - \min_{z' \in \cK}\|Z - z'\|^2 = \|Z\|^2 - \min_{h \in \Reals^k}\|Z - T h\|^2.
\end{align*}
Suppose that $\cZ$ is the unit ball of $\Reals^d$. Then we can restrict the minimization above to $\cH = \{ h \in \Reals^k : \| h \| \le 1\}$. Thus, as already observed by Maurer and Pontil \cite{maurer_14} PCA is equivalent to the task
\begin{align*}
	\min_{f\in \cF_{\text{iso}}}\E_P\left[\min_{h\in\Reals^k:\, \|h\| \le 1} \|Z - f(h)\|^2\right],
\end{align*}
where $\cF_{\text{iso}}$ denotes the family of linear isometries $\Reals^k \to \Reals^d$.

\paragraph{Neural nets} Let $\sigma : \Reals^d \to \Reals^d$ be a fixed nonlinearity. Let $\cF_{\text{nn}}$ consist of all mappings $f: \Reals^k \to \Reals^d$ of the form
\begin{align}\label{eq:nn}
	f(h) = \sum^m_{i=1} c_i \sigma(A_i h + b_i),
\end{align}
where $m \in {\mathbb N}$, $c_i$ are arbitrary real coefficients, $A_i \in \Reals^{d \times k}$ are arbitrary matrices of connection weights, and $b_i \in \Reals^d$ are arbitrary vectors of biases. We can take $\cH$ to be, for example, the $\ell_2$ unit ball in $\Reals^k$, in which case the coding problem consists in finding a vector $h \in \Reals^k$ with $\|h\| \le 1$, such that $Z$ can be best approximated in $L^2(P)$ by passing $h$ through a nonlinear map of the form \eqref{eq:nn}. The class $\cF_{\text{nn}}$ corresponds to neural nets with one hidden layer; we will consider multilayer neural nets in the sequel. This class of coding schemes is closely related to the recent work of Bojanowski et al.~\cite{bojanowski_18} on \textit{generative latent optimization} (GLO), where the aim is to minimize the expected reconstruction error \cref{eq:representation_risk} over the class $\cF$ consisting of multilayer neural nets. Thus, the framework of finite-dimensional coding schemes is sufficiently broad to cover a variety of schemes for latent generative modeling, including \textit{generative adversarial nets} (GAN) \cite{goodfellowetal_16}. Indeed, as shown in \cite{bojanowski_18}, the GLO framework  enables the training of a generator without the need to train the discriminator (which is usually a computational bottleneck), while the learned generator inherits many desirable properties of ordinary GANs, such as arithmetic operations on the representation space or sharpness of generated images.

\subsection{Some results on the expected reconstruction error}

The expected reconstruction error \cref{eq:representation_risk} can be connected to the theory of optimal transport \cite{villani_03} and to rate-distortion theory \cite{berger_71}. While the primary objective of this paper is to study the \textit{learning} of coding schemes, not the (minimum) expected reconstruction error itself, we briefly discuss the ideas and implications below.

\subsubsection{Connection to optimal transport}
\label{sssec:transport}

Using ideas from the theory of optimal transport \cite{villani_03}, we can characterize the expected reconstruction error of a given $f \in \cF$ as the \textit{minimum approximation error} of the data-generating distribution $P$ by probability distributions on $\Reals^d$ that can be realized as pushforwards of probability measures supported on the codebook $\cH$ by the reconstruction map $f$. Before a formal presentation of the result, we introduce the notions from the optimal transport theory: Let $\cP(\cZ)$ be the space of all Borel probability measures on $\cZ$, and let $\cP_p(\cZ)$ with $p \in [1,\infty)$ be the space of all $P \in \cP(\cZ)$ with finite $p$th moment, i.e., $\int_{\cZ} \|z\|^p P(\d z) < +\infty$. Then, we can define \textit{$p$-Wasserstein distance} on $\cP_p(\cZ)$ as
\begin{align*}
	W_p(P,Q) \deq \inf_{M(\cdot\times\cZ) = P \atop M(\cZ \times \cdot) = Q} \left( \E_M\|Z - Z'\|^p \right)^{\frac{1}{p}},
\end{align*}
where the infimum is taken over all couplings of $P$ and $Q$, i.e. probability measures on the product space $\cZ \times \cZ$ with the given marginals $P$ and $Q$. The name ``optimal transport'' comes from the fact that $W^p_p(P,Q)$ can be interpreted as the minimum cost of transporting a unit amount of some material initially distributed as $P$ to a final distribution $Q$, when the unit cost of transporting the material from location $z$ to another location $z'$ is $\|z-z'\|^p$.

Now consider the following recipe for generating a random element of $\Reals^d$: Fix a probability distribution $\pi$ on the codebook $\cH$ and select a measurable map $f : \cH \to \Reals^d$. Then, draw a random element $H \sim \pi$ and pass it through $f$. The probability law of $f(H)$ is called the \textit{pushforward of $\pi$ by $f$} and denoted by $f_\sharp \pi$:  for any Borel set $A \subseteq \Reals^d$,
\begin{align*}
	f_\sharp \pi(A) \deq \pi(f^{-1}[A]),
\end{align*}
where $f^{-1}[A]$ is the preimage of $A$ under $f$.  Then, we have the following result.
\begin{proposition}\label{prop:approx_wass}
	Suppose that $\cZ$ is a compact subset of $\Reals^d$. Then, for any Borel decoder $f: \cH \to \cZ$,
	\begin{align*}
		\risk(P,f) = \inf_{\pi \in \cP(\cH)} W_2^2(P,f_\sharp \pi).
	\end{align*}
Consequently, for any admissible class $\cF$ of reconstruction maps,
\begin{align*} 
	\risk(P,\cF) = \inf_{Q \in \cF_\sharp\cP(\cH)} W^2_2(P,Q),
\end{align*}
where $\cF_\sharp\cP(\cH)$ is the set of all Borel probability measures $Q$ on $\Reals^d$ that can be implemented as a pushforward $f_\sharp \pi$ of some $\pi \in \cP(\cH)$ by some $f \in \cF$.
\end{proposition}
\begin{remark} The assumption that $\cH$ is compact is introduced mainly for the sake of simplicity, and the result may be easily extended to arbitrary Borel sets $\cH$ under appropriate moment conditions on $f_\sharp \pi$.
\end{remark}

It is useful to compare the above proposition to the following classic result of Pollard \cite{pollard_82}: Given a Borel probability measure $P \in \cP_2(\Reals^d)$, let $e_k(P)$ denote the minimum expected reconstruction error for $Z \sim P$ over all $k$-point vector quantizers:
\begin{align}
	e_k(P) \deq \inf_{\cC \subset \Reals^d: |\cC| = k} \E_P\left[\min_{u \in \cC} \|Z-u\|^2\right].
\end{align}
Let $\cP^{(k)} \subset \cP_2(\Reals^d)$ denote the collection of all probability measures  supported by (at most) $k$ points in $\Reals^d$. Then
\begin{align}\label{eq:Pollard}
	e_k(P) = \inf_{Q \in \cP^{(k)}} W^2_2(P,Q).
\end{align}
Recalling the example of vector quantization from \cref{sec:framework}, take $\cH = \{e_1,\ldots,e_k\}$ (the canonical orthonormal basis in $\Reals^d$) and let $\cF$ be the collection of all linear maps $f : \Reals^k \to \Reals^d$. Then any $Q \in \cP^{(k)}$ supported on the set $\{u_1,\ldots,u_k\}$ can evidently be realized as $f_\sharp\pi$ with $\pi(\{e_j\}) = Q(\{u_j\})$ and $f(e_j) = u_j$, $1 \le j \le k$. Since we can now rewrite \eqref{eq:Pollard} as
\begin{align}
	e_k(P) = \inf_{Q \in \cF_\sharp \cP(\cH)} W^2_2(P,Q) \equiv \risk(P,\cF),
\end{align}
one can view Pollard's result \eqref{eq:Pollard} as a special case of \cref{prop:approx_wass}, which allows infinite codebooks and nonlinear reconstruction maps.

This Wasserstein distance characterization of the expected reconstruction error enables an alternative approach to study the generalization error in learning coding schemes. In particular, we can show that the expected reconstruction error with respect to the \textit{empirical distribution} $P_n$ converges to the expected reconstruction error with respect to the data-generating distribution $P$ using the convergence properties of the empirical measure in Wasserstein distance. The idea will be formalized in \cref{ssec:gen_wass}.

\subsubsection{Connection to rate-distortion theory}

For a codebook $\cH$ with finite cardinality, the minimum reconstruction error $\risk(P,\cF)$ can be lower-bounded in terms of information-theoretic quantities originating in rate-distortion theory \cite{berger_71}. We begin by introducing the necessary information-theoretic notions \cite{coveretal_12}: for any two probability measures $\mu, \nu$ on $\Reals^m$, the \textit{Kullback-Leibler divergence} (or \textit{relative entropy})  is defined as
$$
D(\mu\|\nu) \deq \int \d \mu \log \frac{\d \mu}{\d \nu} ,
$$
if $\mu$ is absolutely continuous with respect to $\mu$, where $\frac{\d \mu}{\d \nu}$ is the Radon-Nikodym derivative, and $D(\mu\|\nu) = \infty$ otherwise. The (Shannon) \textit{mutual information} between two random vectors $Z_1, Z_2$ is defined as
\begin{align*}
	I(Z_1;Z_2) \deq D\left(P_{Z_1 Z_2}\| P_{Z_1} \otimes P_{Z_2}\right),
\end{align*}
where $P_{Z_1}, P_{Z_2}, P_{Z_1 Z_2}$ denote the marginal distributions and the joint distribution of $Z_1, Z_2$, respectively. Now, the (information) \textit{distortion-rate function} \cite{bergergibson_98}, with respect to the squared error, is defined as
\begin{align}
	{\mathbb D}(R,P) \deq \inf_{P_{\wh{Z}|Z}: I(Z;\wh{Z}) \leq R} \E\|Z - \wh{Z}\|^2 \label{eq:dr}.
\end{align}
The quantity \cref{eq:dr} arises as a minimum achievable average squared error among any possible \textit{lossy source coding} schemes, i.e. compression/decompression of an analog signal distributed as $P$ using $R$ nats (unit of information corresponding to the natural logarithm). We can now bound the minimum reconstruction error by the distortion-rate function:
\begin{proposition}\label{prop:rd}
	Suppose that the codebook $\cH$ has finite cardinality $k$. Then, for any class of reconstruction maps $\cF$ and any data-generating distribution $P$, we have
	\begin{align*}
		\risk(P,\cF) &\geq {\mathbb D}(\log k, P).
	\end{align*}
\end{proposition}
Expressing the minimum reconstruction error in terms of the distortion-rate function has several advantages. First, the optimization problem \cref{eq:dr} specifying the lower bound is a convex program, and thus can be efficiently approximated (see, e.g., \cite{blahut_72}). Second,  we can estimate ${\mathbb D}(\log k, P)$ from below using the Shannon lower bound \cite{gyorgyetal_99} and get the lower bound $\risk(P,\cF) \succeq \cO(k^{-2/d})$, while the results from high-resolution vector quantization theory \cite{gershogray_12} provide a matching upper bound as $k \to \infty$.

Also note that \cref{prop:rd} can be extended to the case of continuous codebooks via a simple covering number argument, to provide a (possibly loose yet simple) lower bound on the minimum reconstruction risk. For example, suppose that the decoders in $\cF$ are $L$-Lipschitz. Let $\cC_\eps$ be any finite $\eps$-cover of the codebook $\cH$, i.e.,
\begin{align}\label{eq:H_cover}
\sup_{h \in \cH}\min_{c \in \cC_\eps} \| h - c \| \le \eps.
\end{align}
Then, for any $z \in \cZ$, $h \in \cH$, $c \in \cC_\eps$ and any $\lambda \in (0,1)$, we have
\begin{align*}
	\|z-f(c)\|^2 &\le \frac{1}{\lambda} \|z-f(h)\|^2 + \frac{1}{1-\lambda}\|f(c)-f(h)\|^2 \\
	&\le \frac{1}{\lambda}\|z-f(h)\|^2 + \frac{1}{1-\lambda} L^2 \|c-h\|^2,
\end{align*}
where we have used Jensen's inequality and the Lipschitz continuity of $f$. Minimizing both sides over $c \in \cC_\eps$ and $h \in \cH$ and using \eqref{eq:H_cover}, we obtain
\begin{align*}
	\min_{c \in \cC_\eps}\|z - f(c)\|^2 \leq \frac{1}{\lambda}\cdot \min_{h \in \cH}\|z - f(h)\|^2 + \frac{1}{1-\lambda} L^2\eps^2.
\end{align*}
This leads to a lower bound of $\lambda \cdot {\mathbb D}(\log |\cC_\eps|, P) - \frac{\lambda}{1-\lambda}L^2 \eps^2$ on the minimum reconstruction error, for any choice of $\eps > 0$ and $\lambda \in (0,1)$.

\section{Learning coding schemes}\label{sec:learn}
We now consider the problem of \textit{unsupervised learning} of a coding scheme in the situation when the data-generating distribution $P$ is unknown, but we have an access to training samples $Z_1, \ldots, Z_n$ drawn independently from $P$. In particular, we study the \textit{generalization error} with respect to a class $\cF$ of reconstruction maps:
\begin{align}
	\gen(P,\cF) &\deq \sup_{f \in \cF} \left| \risk(P,f) - \risk(P_n, f)  \right|\label{eq:gen_error}\\
	&= \sup_{f \in \cF} \left| \E_P\min_{h \in \cH} \|Z - f(h)\|^2 - \frac{1}{n}\sum_{i=1}^n \min_{h \in \cH}\|Z_i - f(h)\|^2 \right|,\nonumber
\end{align}
where $P_n$ is the \textit{empirical distribution} of the samples, i.e., $P_n(A) = \frac{1}{n}\sum_{i=1}^n \mathbf{1}\{Z_i \in A\}$ for any Borel set $A \subseteq \Reals^d$. In other words, the generalization error measures how accurately the empirical reconstruction error (i.e., the reconstruction error for the training data) approximates the \textit{true} reconstruction error for the data-generating distribution $P$. For simplicity, we drop $P$ and simply write $\gen(\cF)$ when the data-generating distribution is clear from the context.

We remind the reader that any upper bound on the generalization error $\gen(P,\cF)$, e.g., one that holds in expectation or with high probability, provides a theoretical performance guarantee for unsupervised learning using \textit{empirical risk minimization} (ERM):
\begin{align}
	\wh{f} \deq \argmin_{f \in \cF} \risk(P_n,f) = \argmin_{f \in \cF} \sum_{i=1}^n \min_{h \in \cH} \|Z_i - f(h)\|^2. \label{eq:ERM}
\end{align}
Suppose, for simplicity, that a minimizing $\wh{f}$ exists (otherwise, we can consider $\eps$-minimizers and then take $\eps \to 0$). Likewise, assume that there exists some $f^* \in \cF$ that achieves $\risk(P,\cF)$. Then, using the fact that $\risk(P_n,\wh{f}) \le \risk(P_n,f^*)$ by the construction of $\wh{f}$, we have
\begin{align*}
	\risk(P,\wh{f}) - \risk(P,\cF) &= \risk(P,\wh{f}) - \risk(P,f^*) \\
	&= \risk(P,\wh{f}) - \risk(P_n,\wh{f}) + \risk(P_n,\wh{f}) - \risk(P_n,f^*) + \risk(P_n,f^*) - \risk(P,f^*) \\
	&\le 2 \sup_{f \in \cF} |\risk(P_n,f)-\risk(P,f)|.
\end{align*}

In the setting when the representation space $\cH$ is finite, the problem of learning a coding scheme from data and the corresponding generalization error \cref{eq:gen_error} have been studied extensively in the literature on vector quantization and $k$-means clustering \cite{pollard_82, bartlettetal_98, biauetal_08}. The problem of learning a coding scheme with $\cH$ being a compact subset of $\Reals^k$ was addressed first by Maurer and Pontil \cite{maurerpontil_10}, with subsequent work of Vainsencher et al.~\cite{vainsencheretal_11} on dictionary learning, where $\cH$ was the unit sphere in $\Reals^k$ and various sparsity constraints were imposed on the admissible linear reconstruction maps. Related work by Mehta and Gray \cite{mehtagray_13} analyzed the generalization error in the context of predictive sparse coding. In all these works, \textit{linearity} of the reconstruction maps remained the central assumption. One notable exception is the recent work of Mazumdar and Rawat \cite{mazumdar_18}, where the reconstruction maps are taken to be single-layer neural nets with ReLU activation functions. In that work, however, the focus is on approximate recovery (in the Frobenius norm) of the matrix product $AH$, where $A$ is a $m \times k$ matrix of neural network weights and $H$ is the $k \times n$ representation matrix for the $n$ observations $Z_1,\ldots,Z_n$, i.e., the $i$th column of $H$ is the element of $\cH$ corresponding to $Z_i$. However, the problem formulation in \cite{mazumdar_18} does not assume a data-generating distribution $P$ and cannot be interpreted in the form of \cref{eq:gen_error}.

\subsection{A generalization bound in terms of Wasserstein convergence}\label{ssec:gen_wass}

In this section, we show that, as the number of samples $n$ increases, the generalization error $\gen(P,\cF)$ converges to zero with high probability for \textit{any} class $\cF$ of admissible reconstruction maps. More specifically, we have the following result:
\begin{theorem} \label{thm:learn_wasserstein}
	Let $P$ be a probability measure supported on a bounded set $\cZ \subset \Reals^d$ for $d \geq 3$. Then for any $q > 2$ there exists a constant $C_{q,d}$, such that, for any class $\cF$ of admissible reconstruction maps $f : \cH \to \cZ$ and any $\delta \in (0,1)$,
	\begin{align*}
		\gen(P,\cF) \leq C_{q,d}\cdot \diam(\cZ)  \left(\int_{\cZ} \|z\|^q P(\d z)\right)^{\frac{1}{q}} n^{-\frac{1}{d}} +\diam^2(\cZ) \sqrt{\frac{2\log (1/\delta)}{n}}
	\end{align*}
	with probability at least $1-\delta$.
\end{theorem}
\begin{remark} The constant $C_{q,d}$ is related to the so-called \textit{Pierce constant} \cite{dereichetal_13} that appears in the context of high-resolution vector quantization, and is given explicitly in the proof.
\end{remark}

This `umbrella' generalization bound, remarkably, implies that ERM \cref{eq:ERM} is a \textit{universal} representation learning algorithm. In other words, the map $\wh{f}$ computed by ERM achieves nearly the minimum expected reconstruction error with an arbitrary precision and arbitrarily high probability (provided the number $n$ of training instances is large enough) regardless of the choice of $P$ and $\cF$. This property is not true for a general statistical learning problem, where the celebrated ``no-free-lunch'' theorem \cite{sss_14} precludes the existence of such universal learners, and one needs to rely on the finiteness of the hypothesis space capacity \cite{vapnik_98} or on stability assumptions \cite{shalevshwartzetal_10} to show the PAC learnability of the problem.

\cref{thm:learn_wasserstein} has already been partially foreshadowed by the characterization of the reconstruction error in terms of the Wasserstein distance (\cref{prop:approx_wass}). Indeed, for any 
Borel reconstruction map $f: \cH \to \cZ$, we have the following estimate: 
\begin{align*}
	\sup_{\pi}|W_2^2(P,f_\sharp \pi) - W_2^2(P_n, f_\sharp \pi)| &\leq 2\,\diam(\cZ) \cdot \sup_{\pi} \left|W_2(P,f_\sharp \pi) - W_2(P_n,f_\sharp \pi)\right|\\
	&\leq 2\, \diam(\cZ) \cdot W_2(P,P_n),
\end{align*}
where the first inequality uses the identity $a^2 - b^2 = (a-b)(a+b)$, while the second inequality is by the triangle inequality.  Thus, the generalization error can be controlled by the $2$-Wasserstein distance between the empirical distribution $P_n$ and the true distribution $P$; the actual proof, however, goes through the $1$-Wasserstein distance for a more refined bound.

As \cref{thm:learn_wasserstein} relies on the $W_1$ convergence of $P_n$ to $P$, the rate of $n^{-1/d}$ can be improved if we impose additional restrictions on the data-generating distribution $P$. For example, if the upper Wasserstein dimension $d^*_1(P)$ \cite{weedbach_17} is smaller than $d$ (e.g., if $P$ is supported on a lower-dimensional submanifold of $\Reals^d$), then the asymptotic dependency of the bound can be improved to $n^{-1/d^*_1(P)}$. Also note that the convergence in Wasserstein distance (and the generalization bound) can also take place when $\cZ$ is a subset of an infinite-dimensional Hilbert space under suitable assumptions on the moments of $P$; see, e.g., \cite{singhetal_18,lei_18}.

\subsection{Generalization error for reconstruction maps with additional structure}\label{ssec:gen_comp}
\cref{thm:learn_wasserstein} shows that empirical risk minimization is asymptotically consistent under minimal regularity assumptions on the class of reconstruction maps $\cF$. However, the bound requires an exponential growth in the number of training samples as the dimensionality of the data space $\cZ$ grows\footnote{In fact, the constant $C_{q,d}$ also grows exponentially in $d$.}. On the other hand, if the complexity of $\cF$ is constrained in some way, it is possible to use the techniques from empirical process theory to show that the generalization error converges to zero at the rate of $n^{-1/2}$ with high probability \cite{vapnik_98,koltchinskii_11}. Indeed, existing generalization guarantees for the problem \cref{eq:gen_error} are of order $n^{-1/2}$. For example, Maurer and Pontil \cite{maurerpontil_10} show that, when $\cF$ is a family of norm-constrained linear maps and $\cH$ is a unit ball in $\Reals^k$, the generalization bound of order $\cO(k^2/\sqrt{n})$ or $\cO(k\sqrt{\log n/n})$ (depending on the type of norm constraints) can be attained. While the expressive capabilities of linear reconstruction maps are limited, the bound is scalable, as it is completely independent of the dimensionality of the data space $\cZ$.

In light of this, we are now going to develop theoretical upper bounds on the generalization error (with polynomial dependence on the dimensionality of $\cZ$) for a class of structured reconstruction maps that are richer than the class of linear decoders. More specifically, we provide a generalization bound proportional to a suitable \textit{complexity measure} of space $\cF$ and with rate $n^{-1/2}$. The complexity measure adopted in this work is related to the \textit{entropy integral} \cite{talagrand_14} from the theory of empirical processes. Before presenting it, we need to introduce some definitions first. Let $A$ be a subset of a pseudometric space\footnote{A pseudometric on a set $T$ is a map $d : T \times T \to \Reals_+$ that satisfies the triangle inequality, $d(s,t) \le d(s,t')+d(t',t)$ for all $s,t,t' \in T$, but $d(s,t) = 0$ does not necessarily imply that $s = t$.} $(T,d)$. A finite set $S \subset T$ is an \textit{$\eps$-net} of $A$ if
\begin{align*}
	\sup_{t \in A}\min_{s \in S} d(s,t) \le \eps.
\end{align*}
The \textit{$\eps$-covering number} of $A$ is then defined as
\begin{align*}
	\cN(A,d,\eps) \deq \min \left\{ |S| : S \text{ is an $\eps$-net of $A$}\right\}
\end{align*}
With these definitions in place, we take our complexity measure of $\cF$ to be
\begin{align}
	\comp(\cF) \deq \inf_{\alpha \geq 0} \left\{  \frac{\alpha\sqrt{n}}{6} + \int_{\alpha}^{\frac{\diam(\cZ)}{2}} \sqrt{\log\cN(\cF,\|\cdot\|_{\cH},u)}\, \d u \right\}, \label{eq:ent_int}
\end{align}
where $\cN(\cF,\|\cdot\|_\cH,\cdot)$ is the covering number of $\cF$ in the pseudometric 
$$
\|f-f'\|_\cH \deq \sup_{h \in \cH}\|f(h)-f'(h)\|.
$$
The entropy integral \cref{eq:ent_int} can be linked to other complexity measures used in empirical process theory, such as Rademacher and Gaussian complexities, via Dudley's entropy integral methods \cite{dudley_67} and Sudakov minoration \cite{latala_97}. By using the entropy integral as a complexity measure, we can prove the following general result, which will be applied to specific examples of reconstruction maps in \cref{sec:stack}:
\begin{theorem}\label{thm:learn_complexity}
	Let $\cZ \subset \Reals^d$ be a bounded set. Then, for any class $\cF$ of admissible reconstruction maps and any $\delta \in (0,1)$,
	\begin{align*}
		\gen(\cF) \leq \frac{96\, \diam(\cZ)}{\sqrt{n}}  \comp(\cF) + \diam^2(\cZ)\sqrt{\frac{2\log(2/\delta)}{n}}, \quad \text{w.p.}\:\:1-\delta.
	\end{align*}
\end{theorem}

\cref{thm:learn_complexity} extends and refines the bound of Vainsencher et al.~\cite[Lemma 21]{vainsencheretal_11} based on covering numbers. More specifically, \cref{thm:learn_complexity} could be used to provide generalization guarantees for a family of nonlinear reconstruction maps, and the proof incorporates the \textit{chaining} of successively finer covers \cite{talagrand_14} instead of a single covering step, as in \cite{vainsencheretal_11}. This chaining-based bound is particularly useful when one considers a more general class of reconstruction maps than linear maps with a given upper bound on the operator norm. For example, consider the following set-up: Let $\cF$ be a family of $d \times k$ matrices with entrywise $\ell_1$-norm at most $M$, and let $\cH$ be a unit ball in $\Reals^k$. Also, assume that we are using $\ell_2$ norms on both the input and the output spaces. Then, using the empirical method of Maurey (see \cite{zhang_02} and references therein), one can show that the logarithm of the covering number can be bounded as
\begin{align}
	\log\cN(\cF,\|\cdot\|_{\cH},\eps) \leq \log\cN(\cF,\|\cdot\|_2,\eps) \leq \left\lceil\frac{M^2}{\eps^2}\right\rceil\log\left(1+\frac{2dk \eps^2}{M^2}\right), \label{eq:jbm}
\end{align}
where the first inequality holds by the relationship between $\ell_2$-induced operator norm and the entrywise $\ell_2$-norm (which in this case coincides with the Frobenius norm), which we denote by $\|\cdot\|_2$. Combined with \cref{thm:learn_complexity}, this leads to a generalization bound of order $\cO(M\sqrt{\log dk}\log n/\sqrt{n})$. On the other hand, the method based on single-step covering does not provide a bound of the same order for any possible covering radius $\eps$.

\section{Deep neural nets as reconstruction maps} \label{sec:stack}
We now consider a family of nonlinear reconstruction maps constructed by composing multiple layers of nonlinear transformations with a given structure. Such multilayer generative models are commonly used in the domain of autoencoders \cite{vincent_10, ollivier_14} or generative adversarial networks \cite{goodfellowetal_14}, including the case of generative latent optimization (GLO) \cite{bojanowski_18} which uses a generator composed mainly of stacked transposed convolutional layers. Formally, we consider a family of nonlinear maps of the form
\begin{align}
	f_\ell\left(h;A_{1:\ell}\right) \deq F_\ell\big(F_{\ell-1}\big(\cdots F_1(h;A_1)\cdots;A_{\ell-1}  \big);A_\ell\big), \label{eq:stack_dec}
\end{align}
where $\ell \in \mathbb{N}$ is the \textit{depth} (or the \textit{number of layers}). Here $A_{1:\ell} = \{A_1,\ldots,A_\ell\}$ is the collection of the layerwise parameters, and, for each $j \in \{1,\ldots,\ell\}$, $F_j(\cdot;A_j) : \Reals^{w_{j-1}} \to \Reals^{w_j}$ is a nonlinear map parametrized by $A_j$. Here, $w_j$ is the \textit{width} of the $j$th layer, and we take $w_0 = k$ (the input dimension) and $w_\ell = d$ (the output dimension). The family of all depth-$\ell$ reconstruction maps is then defined as
\begin{align}
	\cF_\ell \deq \left\{ \pi_\cZ \circ f_\ell\left( \cdot ;A_{1:\ell}\right)~\Big|~ A_j \in \cA_j, \quad \forall j \in \{1,\ldots,\ell\} ~\right\}, \label{eq:stack_dec_fam}
\end{align}
where $\cA_1,\ldots,\cA_\ell$ are a fixed family of layerwise parameter sets, and
\begin{align}
	\pi_\cZ(\xi) \deq \argmin_{z \in \cZ} \|\xi-z\|
\end{align}
is the projection onto $\cZ$.
Generalization bounds involving such `deep' neural networks have been studied extensively in the context of \textit{supervised learning}, where one is given $n$ i.i.d.\ samples $(X_1,Y_1),\ldots,(X_n,Y_n)$, and the objective is to learn the parameters $\wh{A}_{1:\ell}$ of a neural net $\wh{f}$, such that $\wh{Y} = \wh{f}(X)$ is an accurate prediction of $Y$, and the generalization error is given by
\begin{align}\label{eq:nn_prediction}
\sup_{A_{1:\ell} \in \cA_{1:\ell}} \left|\frac{1}{n}\sum^n_{i=1}\|Y_i - f_\ell(X_i; A_{1:\ell})\|^2 - \E\|Y-f_\ell(X; A_{1:\ell})\|^2\right|.
\end{align}
One of the classical results in this direction is the work of Anthony and Bartlett \cite{anthony_09}, which provides upper bounds on the Rademacher averages of neural network predictors via the VC dimension. More recent works focus on providing \textit{scalable} generalization bounds with weaker dependencies on the depth and width (dimensionality of layerwise outputs) of neural nets as an attempt to explain the empirically observed ability of neural nets to generalize well. In these works, Rademacher averages of neural nets are bounded via the contraction principle \cite{neyshaburetal_15, golowichetal_17}, covering number arguments \cite{bartlettetal_17, lietal_18}, or approximations by simpler classes of functions \cite{golowichetal_17, barronetal_18}. By contrast, the problem of learning a $k$-dimensional representation with neural nets as reconstruction maps is an unsupervised learning problem, and its analysis involves the supremum
\begin{align}
	&\sup_{f \in \cF_\ell} \left|\frac{1}{n}\sum^n_{i=1}e_f(Z_i) - \E_P[e_f(Z)]\right|\label{eq:nn_reconstruction} \\
	& \qquad = \sup_{A_{1:\ell} \in \cA_{1:\ell}} \left|\frac{1}{n}\sum^n_{i=1}\min_{h \in \cH} \|Z_i - \pi_\cZ (f_\ell(h;A_{1:\ell}))\|^2 - \E_P\left[\min_{h \in \cH}\|Z - \pi_\cZ(f(h;A_{1:\ell}))\|^2\right]\right|. \nonumber
\end{align}
Most of the ideas used in the analysis of \eqref{eq:nn_prediction}, with the exception of covering number results via \cref{thm:learn_complexity}, cannot be employed directly for the analysis of \eqref{eq:nn_reconstruction}, as the terms of the form $\min_{h\in\cH}\|Z_i - \pi_\cZ \circ f(h)\|^2$ preclude the efficient `peeling off' \cite{neyshaburetal_15} of neural network layers.

\subsection{Fully connected neural nets} \label{ssec:dense}
We first consider the simplest scenario of fully connected (or \textit{dense}) neural nets, which is one of the elementary building blocks of deep neural architectures. In layer $j$, each neuron calculates a weighted sum of the outputs from all the neurons in layer $j-1$ and passes it through a nonlinearity $\sigma_j : \Reals^{w_j} \to \Reals^{w_j}$ (referred to as the activation function). The layerwise operation of fully connected neural networks can be described as
\begin{align}
	F_j(\xi;A) = \sigma_j(A\xi),
\end{align}
where the parameter of $F_j$ is the \textit{weight matrix} $A \in \Reals^{w_{j} \times w_{j-1}}$, with $A_{ik}$ denoting the connection weight from the $k$th neuron in the $(j-1)$th layer to the $i$th neuron in the $j$th layer. We assume that the weight matrix $A$ lies in the parameter space $\cA_j$ with entrywise $\ell_1$-norm constraints, i.e.,
\begin{align*}
	\cA_j \subseteq \left\{ A~|~A\in\Reals^{w_j \times w_{j-1}}, \|A\|_1 \leq M_j\right\}, \qquad j = 1,\ldots,\ell
\end{align*}
for some $M_1, \ldots, M_\ell > 0$. Each activation function $\sigma_j(\cdot)$ is assumed to be $L_j$-Lipschitz (with respect to $\ell_2$-norm on both the input and the output) and to have the zero-in zero-out (ZIZO) property, i.e., $\sigma_j(\mathbf{0}) = \mathbf{0}$. Examples of such activation functions include the rectified linear unit (ReLU), which applies the map $u \mapsto u{\bf 1}_{\{u \ge 0\}}$ componentwise, the leaky ReLU, which applies the map $u \mapsto u{\bf 1}_{\{u \ge 0\}} + \delta u {\bf 1}_{\{u < 0\}}$ for some small $\delta >0$, and the hyperbolic tangent activation function that applies the map $u \mapsto \tanh u$ componentwise. An example of an activation function that does not have the ZIZO property is the \textit{sigmoid} activation function that applies the map $u \mapsto \frac{1}{1+e^{-u}}$ componentwise; we will discuss the generalization bounds for neural net reconstruction maps with such activation functions in \cref{ssec:sigmoid}.

\sloppypar We now present the following generalization bound, which can be thought of as a representation-learning counterpart of \cite{bartlettetal_17}.
\begin{theorem}[wide net] \label{thm:dense} Let $\cZ$ be a compact convex subset of $\Reals^d$ containing $\mathbf{0}$.	Under the above assumptions, for any $\delta \in (0,1)$,
	\begin{align*}
		\gen(\cF_\ell) &\leq \frac{48 \sqrt{2} \cdot \ell \log n}{\sqrt{n}}\cdot \diam(\cZ) \|\cH\|_\infty \left(\prod_{j=1}^\ell L_j M_j \right)\sqrt{\sum_{j=1}^\ell\log (2w_{j-1}w_j + 1)}\\
		&\qquad+ \frac{8\,\diam^2(\cZ)}{\sqrt{n}} + \diam^2(\cZ)\sqrt{\frac{2\log(2/\delta)}{n}}
	\end{align*}
with probability at least $1-\delta$. Here, $\|\cH\|_\infty \deq \sup_{h \in \cH}\|h\|$.
\end{theorem}
If $L_jM_j \le 1$ for each $j$, this bound is of order $\tilde\cO(\sqrt{\ell^3/n})$, and is only logarithmically dependent on the \textit{width} of the neural network, as are the state-of-the-art bounds \cite{bartlettetal_17, barronetal_18} on the generalization error for supervised learning using neural nets.

On the other hand, there are two scenarios where \cref{thm:dense} falls short of being optimal, as we will see in \cref{thm:dense_deep} below. First, this depth dependence of $\ell^{3/2}$ is not optimal in general, if one is willing to sacrifice in terms of width-dependency. Indeed, the number of parameters in the whole network is $\sum_{i=1}^\ell w_{i-1}w_i$, which implies that the optimal dependence on depth may be of order $\ell^{1/2}$ for small width. Second, the multiplicative term  $\|\cH\|_\infty (\prod_{j=1}^\ell L_j M_j)$, which is a Lipschitz constant for the composite reconstruction map $f_\ell$, is excessively large in the case where the diameter of data space $\diam(\cZ)$ is small in comparison. In the case of supervised learning with neural nets, Barron and Klusowski \cite{barronetal_18} recently showed that one can replace the product-of-norm constant with norm-of-product, via sparsification methods combined with a technique specifically developed for the ReLU activation functions. For the problem of learning a coding scheme, however, it turns out that one can easily replace the constant $\|\cH\|_\infty (\prod_{j=1}^\ell L_j M_i)$ with $\diam(\cZ)$. 

The following generalization bound, based on the volumetric estimate for the covering numbers and the one-step approximation argument of Vainsencher et al.~\cite{vainsencheretal_11}, complements \cref{thm:dense} in the above two aspects.
\begin{theorem}[deep net] \label{thm:dense_deep}
	Suppose that $\cA_j$ is a family of matrices with spectral norms at most $M_j$, instead of $\ell_1$-norm, for each $j \in \{1,\ldots,\ell\}$. Then, for any $\delta \in (0,1)$:
	\begin{align*}
		\gen(\cF_\ell) &\leq \diam^2(\cZ)\frac{\sqrt{2\sum_{j=1}^\ell w_{j-1}w_j}}{\sqrt{n}}\sqrt{\log \left(3\ell\sqrt{n}\|\cH\|_\infty\left(\prod_{j=1}^\ell L_jM_j\right)\right)}\\
		&\qquad+ \diam^2(\cZ)\sqrt{\frac{2\log (2/\delta)}{n}} + \frac{4}{\sqrt{n}}, \qquad \text{w.p. } 1-\delta.
	\end{align*}
\end{theorem}
Note that the $\ell_1$-norm constraint on the weight matrices automatically implies that the spectral norm of the weight matrices are bounded from above by same constant.

\subsection{Convolutional neural nets} \label{ssec:convnet}
As \cref{thm:dense_deep} implies, the generalization error can be upper bounded by the term proportional to the square root of the number of parameters, even when the neurons are not fully connected and the effective number of parameters is strictly smaller than $\sum_{j=1}^\ell w_{j-1}w_j$. One important example of this is the case of convolutional neural networks (also referred to as CNNs or ConvNets) \cite{lecunetal_98,goodfellowetal_16}, which are widely used in the context of image data. Rather than calculating the full inner product of the inputs and the weights, each neuron in a convolutional layer takes the inner product of a limited number of outputs from the \textit{spatially close} neurons in the previous layer and the \textit{filter weights} of the convolution filter, which are shared among all neurons. Often, more than one channel of convolution filters is used; the outputs of such a layer will be equipped with an \textit{internal depth} equal to the number of filters being used. The layerwise operation has the form
\begin{align}
	F_j\left(\xi;A^{(1:v_j)}\right) = \sigma\left(\conv\left(\xi;A^{(1:v_j)}\right)\right), \label{eq:convolutional_layer}
\end{align}
for some convolution operator $\conv$ (specified below), input $\xi$, convolution filters $A^{(1:v_j)}$, and an activation function $\sigma$. For simplicity, we assume that all the $\sigma$ are $1$-Lipschitz with respect to $\ell_2$-norm and have the ZIZO property.

First, we consider the simplest case of one-dimensional convolutions, where the input $\xi$ to the $j$th layer is a $w_{j-1} \times v_{j-1}$ matrix, and each of the $v_j$ convolution filters $A^{(1)},\ldots,A^{(v_j)}$ is a $u_j \times v_{j-1}$ matrix for some \textit{filter width} $u_j$. The convolution operation $\xi \mapsto \conv\left(\xi;A^{(1:v_j)}\right) \in \Reals^{w_j \times v_j}$ is then specified by
\begin{align}
	\left( \conv\left(\xi;A^{(1:v_j)}\right) \right)_{i,k} = \sum_{i' = 1}^{u_j} \sum_{j' = 1}^{v_{j-1}} A_{i',j'}^{(k)} \xi_{i'+s(i-1)+\frac{1-u_j}{2}, j'},  \qquad i = 1,\ldots,w_j;\, k = 1,\ldots,v_j \label{eq:convolution_1d}
\end{align}
for some \textit{stride} $s$ denoting the scale of the convolution filter shift for each output entry. Note that we are using the convention $\xi_{i,k} = 0$ when $i \notin \{1,\ldots,w_{j-1}\}$ or $k \notin \{1,\ldots,v_{j-1}\}$. We also assume that the convolution filters have constrained $\ell_1$-norms in the following sense:
\begin{align*}
	\cA_j \subseteq \left\{A^{(1:v_j)}~|~A^{(k)} \in \Reals^{u_j \times v_{j-1}}, \sqrt{\sum_{k=1}^{v_j} \|A^{(k)}\|_1^2} \leq M_j\right\}
\end{align*}
for some constants $M_1,\ldots,M_\ell > 0$. Then, we can prove the following result.
\begin{theorem}[Spatial dimension $1$]\label{thm:cnn_1d}
	Under the above assumptions, for any $\delta \in (0,1)$:
	\begin{align*}
		\gen(\cF_\ell) &\leq \diam^2(\cZ) \frac{\sqrt{2\sum_{j=1}^{\ell} u_j v_{j-1}v_j}}{\sqrt{n}}\sqrt{\log \left(3\ell\sqrt{n}\|\cH\|_{\infty}\left(\prod_{j=1}^\ell M_j\right)\right)}\\
		&\qquad +\diam^2(\cZ)\sqrt{\frac{2\log(2/\delta)}{n}} + \frac{4}{\sqrt{n}} \qquad \text{w.p. } 1-\delta.
	\end{align*}
\end{theorem}
Notice that the generalization bound in \cref{thm:cnn_1d} is now proportional to the square root of  $\sum_{j=1}^\ell u_j v_{j-1}v_j$, which is the number of parameters in the filter matrices of all layers, and which is strictly smaller than the total number $\sum_{j=1}^\ell w_{j-1}v_{j-1}w_jv_j$ of all possible neural connections. The proof of \cref{thm:cnn_1d} relies on the following variant of Young's convolution inequality:
\begin{align}
	\|\conv(\xi;A^{(1:v)})\|_2 \leq \sqrt{\sum_{k=1}^v \|A^{(k)}\|_1^2}\cdot \|\xi\|_2. \label{eq:young_ineq}
\end{align}
The inequality \cref{eq:young_ineq} enables the use of the usual `peeling-off' machinery used for the analysis of fully connected neural networks (see the proof of \cref{thm:cnn_1d} in the appendices for details and the proof of \cref{eq:young_ineq}).

Comparing with the generalization error bounds for supervised learning in the recent work of Li et al.\ \cite{lietal_18} that analyzes convolutional neural nets for prediction, we emphasize two key differences. First, our method does not require orthogonality of the convolutional filters, and can be applied to an arbitrary collection of norm-constrained matrices. Second, as the convolution inequality \cref{eq:young_ineq} can be extended naturally to higher-order tensors, the generalization bound can be provided for the cases of higher spatial dimensions, e.g., images (spatial dim.\ $2$) or videos (spatial dim.\ $3$).

To formalize the second point, consider the following setup: For the $j$th layer, the input $\xi$ takes the form of a tensor of order $m+1$ for some spatial dimension $m$, i.e., $\xi \in \Reals^{w_{j-1,1} \times \cdots \times w_{j-1,m} \times v_{j-1}}$. The parameter space $\cA_j$ is composed of $v_j$ channels of weight tensors $A^{(k)}$ of dimension $u_{j,1} \times \cdots \times u_{j,m} \times v_{j-1}$ with some filter width $u_{j,1}, \ldots, u_{j,m}$, and with a norm constraint
\begin{align*}
	\cA_j \subseteq \left\{A^{(1:v_j)}~\Bigg|~ \sqrt{\sum_{k=1}^{v_j} \|A^{(k)}\|_1^2} \leq M_j\right\}.
\end{align*}
Given strides $s_{j,1}, \ldots, s_{i,m}$ for each spatial dimension, the convolution can be characterized as
\begin{align}
	\left(\conv\left(\xi;A^{(1:v_j)}\right)\right)_{r_1:r_m, k} &\deq \sum_{r'_1 = 1}^{u_{j,1}} \cdots \sum_{r'_m = 1}^{u_{j,m}} \sum_{j' = 1}^{v_{j-1}} A^{(k)}_{r'_1:r'_m, j'} \nonumber\\
	& \qquad \qquad \cdot \xi_{r'_1 + s_1(r_1-1) + \frac{1-u_{j,1}}{2}, \cdots, r'_m + s_m(r_m-1) + \frac{1-u_{j,m}}{2}, j'}. \label{eq:convolution_ld}
\end{align}
Then, we can prove the following result.
\begin{theorem}[Spatial dimension $l$]\label{thm:cnn_ld}
	Under the above assumptions, for any $\delta \in (0,1)$:
	\begin{align*}
		\gen(\cF_\ell) &\leq \diam^2(\cZ)\frac{\sqrt{2\sum_{i=1}^\ell v_{i-1}v_i(\prod_{j=1}^l u_{i,j})}}{\sqrt{n}}\sqrt{\log\left(3\ell\sqrt{n}\|\cH\|_\infty \left(\prod_{i=1}^\ell M_i\right)\right)}\\
		&\qquad + \diam^2(\cZ)\sqrt{\frac{2\log(2/\delta)}{n}} + \frac{4}{\sqrt{n}} \qquad \text{w.p. } 1-\delta.
	\end{align*}
\end{theorem}
\begin{remark}
	Along with proper shifts, the formula \cref{eq:convolution_ld} is general enough to cover the case of fractional strides (also called transposed convolution or deconvolutional networks) \cite{zeileretal_10}, which is a building block of generative adversarial networks \cite{goodfellowetal_14}, with the convention $\xi_{i,j} = 0$ for non-integer values $i,j$.
\end{remark}
\begin{remark}
	Max-pooling layers, which are commonly inserted between convolutional layers to reduce the dimensionality of the representation \cite[Ch.~9]{goodfellowetal_16}, are $1$-Lipschitz mappings with the ZIZO property. Hence, their presence does not affect the generalization bound.
\end{remark}

\subsection{Nonlinearities without the ZIZO property} \label{ssec:sigmoid}
In \cref{ssec:dense,ssec:convnet}, it was assumed that the activation functions had the zero-in/zero-out (ZIZO) property, i.e., $\sigma(\mathbf{0}) = \mathbf{0}$; in the proof of \cref{thm:dense}, this assumption enables a recursive breakdown of $\sup_{h\in\cH} \|f_\ell(h) - \tilde{f}_\ell(h)\|$ for any two $\ell$-layer neural nets $f_\ell$ and $\tilde{f}_\ell$ into $\ell$ terms, each proportional to the $\ell_1$ distance between the weight matrices. The ZIZO property provides a ready way  to upper-bound the magnitude of the outputs from each layer.

However, there are several commonly used activation functions, e.g., the \textit{sigmoid} ($u \mapsto \frac{1}{1+e^{-u}}$ applied entrywise), which do not satisfy this assumption. On the other hand, the outputs from such activation functions are often uniformly bounded, which opens up an alternative path to control the pseudometric $\sup_{h\in\cH}\|f_\ell(h) - \tilde{f}_\ell(h)\|$.

To formalize the idea, let us revisit the setting of fully connected neural nets as reconstruction maps: we assume that the layerwise operation is given as $F_j(\xi;A) = \sigma_j(A\xi)$, where the weight matrix $A \in \Reals^{w_j \times w_{j-1}}$ has $\ell_1$-norm no greater than $M_j$. In addition, we assume that the activation functions $\sigma_j$ for each layer are $L_j$-Lipschitz (with respect to the $\ell_2$-norm), and that their outputs are bounded in norm by $B_j$, i.e. $\|\sigma_j(x)\| \leq B_j$ for any $x \in \Reals^{w_j}$. Then we can prove the following generalization bound:
\begin{theorem}\label{thm:dense_sigmoid}
	Under the above assumptions, for any $\delta \in (0,1)$:
	\begin{align*}
		\gen(\cF_\ell) &\leq \frac{48\sqrt{2} \log n}{\sqrt{n}}\left(\sum_{i=1}^\ell B_{i-1}\left(\prod_{j=i}^\ell L_j M_j\right)\right)\sqrt{\sum_{i=1}^\ell \log(2 w_{i-1}w_i + 1)}\\
		&\qquad + \frac{8\,\diam^2(\cZ)}{\sqrt{n}} + \diam^2(\cZ)\sqrt{\frac{2\log(2/\delta)}{n}}, \qquad \text{w.p. } 1-\delta.
	\end{align*}
\end{theorem}
Unlike \cref{thm:dense}, which was independent of the width of the neural net up to logarithmic factors, the above generalization bound may grow as the width of the neural net gets larger; for example, the $\ell_2$-norm of the $w$-dimensional vector processed by sigmoid activations can be as large as $\sqrt{w}$, which gives the generalization bound roughly of order $\tilde{O}(\sqrt{\ell^2 w/n})$.

\section*{Acknowledgments}
The authors would like to thank Aolin Xu and Matus Telgarsky for helpful discussions. This work was supported in part by NSF grant nos.\ CIF-1527 388 and CIF-1302438, and in part by the NSF CAREER award 1254041.

\bibliography{learncode.bbl}

\begin{thebibliography}{10}

\bibitem{anthony_09}
Martin Anthony and Peter~L. Bartlett.
\newblock {\em Neural Network Learning: Theoretical Foundations}.
\newblock Cambridge University Press, 1999.

\bibitem{barronetal_18}
A.~R. Barron and J.~Klusowski.
\newblock Approximation and estimation for high-dimensional deep learning
  networks.
\newblock arXiv preprint 1809.03090, September 2018.

\bibitem{bartlettetal_98}
P.~L. Bartlett, T.~Linder, and G.~Lugosi.
\newblock The minimax distortion redundancy in empirical quantizer design.
\newblock {\em IEEE Transactions on Information Theory}, 44(5):1802--1813,
  September 1998.

\bibitem{bartlettetal_17}
Peter~L Bartlett, Dylan~J Foster, and Matus~J Telgarsky.
\newblock Spectrally-normalized margin bounds for neural networks.
\newblock In {\em Advances in Neural Information Processing Systems 30}, pages
  6240--6249, 2017.

\bibitem{bengioetal_13}
Y.~Bengio, A.~Courville, and P.~Vincent.
\newblock Representation learning: A review and new perspectives.
\newblock {\em IEEE Transactions on Pattern Analysis and Machine Intelligence},
  35(8):1798--1828, August 2013.

\bibitem{berger_71}
T.~Berger.
\newblock {\em Rate Distortion Theory: A Mathematical Basis for Data
  Compression}.
\newblock Prentice-Hall electrical engineering series. Prentice-Hall, 1971.

\bibitem{bergergibson_98}
T.~Berger and J.~D. Gibson.
\newblock Lossy source coding.
\newblock {\em IEEE Transactions on Information Theory}, 44(6):2693--2723,
  October 1998.

\bibitem{biauetal_08}
G.~Biau, L.~Devroye, and G.~Lugosi.
\newblock On the performance of clustering in {Hilbert} spaces.
\newblock {\em IEEE Transactions on Information Theory}, 54(2):781--790,
  February 2008.

\bibitem{blahut_72}
R.~Blahut.
\newblock Computation of channel capacity and rate-distortion functions.
\newblock {\em IEEE Transactions on Information Theory}, 18(4):460--473, July
  1972.

\bibitem{bojanowski_18}
Piotr Bojanowski, Armand Joulin, David Lopez-Pas, and Arthur Szlam.
\newblock Optimizing the latent space of generative networks.
\newblock In {\em Proceedings of the 35th International Conference on Machine
  Learning}, volume~80 of {\em Proceedings of Machine Learning Research}, pages
  600--609, Stockholm, Sweden, Jul 2018. PMLR.

\bibitem{boucheronetal_13}
St{\'e}phane Boucheron, G{\'a}bor Lugosi, and Pascal Massart.
\newblock {\em Concentration inequalities: A nonasymptotic theory of
  independence}.
\newblock Oxford university press, 2013.

\bibitem{chafai_12}
Djalil Chafa\"{i}, Olivier Gu{\'e}don, Guillaume Lecu{\'e}, and Alain Pajor.
\newblock {\em Interactions between compressed sensing, random matrices, and
  high dimensional geometry}, volume~37 of {\em Panoramas et Synth{\`e}ses}.
\newblock {Soci{\'e}t{\'e} Math{\'e}matique de France}, December 2012.

\bibitem{coveretal_12}
T.~M. Cover and J.~A. Thomas.
\newblock {\em Elements of information theory}.
\newblock John Wiley \& Sons, 2012.

\bibitem{dereichetal_13}
S.~Dereich, M.~Scheutzow, and R.~Schottstedt.
\newblock Constructive quantization: Approximation by empirical measures.
\newblock {\em Annales de l'institut Henri Poincar\'e, probabilit\'es et
  statistiques}, 49(4):1183--1203, 11 2013.

\bibitem{dudley_67}
R.~M. Dudley.
\newblock The sizes of compact subsets of {H}ilbert space and continuity of
  {G}aussian processes.
\newblock {\em Journal of Functional Analysis}, 1(3):290--330, 1967.

\bibitem{gershogray_12}
A.~Gersho and R.~M. Gray.
\newblock {\em Vector quantization and signal compression}, volume 159 of {\em
  The Springer International Series in Engineering and Computer Science}.
\newblock Springer, 1992.

\bibitem{golowichetal_17}
N.~Golowich, A.~Rakhlin, and O.~Shamir.
\newblock Size-independent sample complexity of neural networks.
\newblock arXiv preprint 1712.06541, December 2017.

\bibitem{goodfellowetal_16}
I.~Goodfellow, Y.~Bengio, and A.~Courville.
\newblock {\em Deep Learning}.
\newblock MIT Press, Cambridge, MA, 2016.

\bibitem{goodfellowetal_14}
I.~Goodfellow, J.~Pouget-Abadie, M.~Mirza, B.~Xu, D.~Warde-Farley, S.~Ozair,
  A.~Courville, and Y.~Bengio.
\newblock Generative adversarial nets.
\newblock In {\em Advances in Neural Information Processing Systems 27}, pages
  2672--2680, 2014.

\bibitem{gyorgyetal_99}
A.~Gyorgy, T.~Linder, and K.~Zeger.
\newblock On the rate-distortion function of random vectors and stationary
  sources with mixed distributions.
\newblock {\em IEEE Transactions on Information Theory}, 45(6):2110--2115,
  September 1999.

\bibitem{jolliffe_02}
Ian~T. Jolliffe.
\newblock {\em Principal Component Anaysis}.
\newblock Springer-Verlag, 2 edition, 2002.

\bibitem{koltchinskii_11}
V.~Koltchinskii.
\newblock {\em Oracle inequalities in empirical risk minimization and sparse
  recovery problems}, volume 2033 of {\em Lecture notes in mathematics}.
\newblock Springer, 2011.

\bibitem{latala_97}
R.~Lata{\l}a.
\newblock Sudakov minoration principle and supremum of some processes.
\newblock {\em Geometric \& Functional Analysis}, 7(5):936--953, 1997.

\bibitem{lecunetal_98}
Y.~LeCun, L.~Bottou, Y.~Bengio, and P.~Haffner.
\newblock Gradient-based learning applied to document recognition.
\newblock {\em Proceedings of the IEEE}, 86(11):2278--2324, November 1998.

\bibitem{lei_18}
J.~Lei.
\newblock Convergence and concentration of empirical measures under
  {Wasserstein} distance in unbounded functional spaces, April 2018.

\bibitem{lietal_18}
X.~Li, J.~Lu, Z.~Wang, J.~Haupt, and T.~Zhao.
\newblock On tighter generalization bound for deep neural networks: {CNNs},
  {ResNets}, and beyond.
\newblock arXiv preprint 1806.05159, June 2018.

\bibitem{linder_02}
T.~Linder.
\newblock Learning-theoretic methods in vector quantization.
\newblock In {\em Principles of nonparametric learning}, pages 163--210.
  Springer, 2002.

\bibitem{maurerpontil_10}
A.~Maurer and M.~Pontil.
\newblock $k$-dimensional coding schemes in {Hilbert} spaces.
\newblock {\em IEEE Transactions on Information Theory}, 56(11):5839--5846,
  November 2010.

\bibitem{maurer_14}
Andreas Maurer.
\newblock A chain rule for the expected suprema of {Gaussian} processes.
\newblock {\em Theoretical Computer Science}, 650:109--122, 2016.
\newblock Algorithmic Learning Theory.

\bibitem{mazumdar_18}
Ayra Mazumdar and Ankit~Singh Rawat.
\newblock Representation learning and recovery in the {ReLU} model.
\newblock arXiv preprint 1803.04304, March 2018.

\bibitem{mehtagray_13}
Nishant Mehta and Alexander Gray.
\newblock Sparsity-based generalization bounds for predictive sparse coding.
\newblock In {\em Proceedings of the 30th International Conference on Machine
  Learning}, volume~28 of {\em Proceedings of Machine Learning Research}, pages
  36--44, Atlanta, Georgia, USA, June 2013.

\bibitem{neyshaburetal_15}
Behnam Neyshabur, Ryota Tomioka, and Nathan Srebro.
\newblock Norm-based capacity control in neural networks.
\newblock In {\em Proceedings of The 28th Conference on Learning Theory},
  volume~40 of {\em Proceedings of Machine Learning Research}, pages
  1376--1401, Paris, France, July 2015.

\bibitem{ollivier_14}
Y.~Ollivier.
\newblock Auto-encoders: reconstruction versus compression.
\newblock arXiv preprint 1403.7752, March 2014.

\bibitem{pollard_82}
D.~Pollard.
\newblock Quantization and the method of $k$-means.
\newblock {\em IEEE Transactions on Information Theory}, 28(2):199--205, March
  1982.

\bibitem{rieder_78}
Ulrich Rieder.
\newblock Measurable selection theorems for optimization problems.
\newblock {\em Manuscripta Mathematica}, 24(1):115--131, March 1978.

\bibitem{sss_14}
Shai Shalev-Shwartz and Shai Ben-David.
\newblock {\em Understanding machine learning: From theory to algorithms}.
\newblock Cambridge university press, 2014.

\bibitem{shalevshwartzetal_10}
Shai Shalev-Shwartz, Ohad Shamir, Nathan Srebro, and Karthik Sridharan.
\newblock Learnability, stability and uniform convergence.
\newblock {\em Journal of Machine Learning Research}, 11:2635--2670, October
  2010.

\bibitem{singhetal_18}
S.~Singh and B.~P{\'o}czos.
\newblock Minimax distribution estimation in {Wasserstein} distance, February
  2018.

\bibitem{talagrand_14}
M.~Talagrand.
\newblock {\em Upper and lower bounds for stochastic processes: modern methods
  and classical problems}.
\newblock Springer, 2014.

\bibitem{vainsencheretal_11}
Daniel Vainsencher, Shie Mannor, and Alfred~M. Bruckstein.
\newblock The sample complexity of dictionary learning.
\newblock {\em Journal of Machine Learning Research}, 12:3259--3281, November
  2011.

\bibitem{vandegeer_00}
Sara van~de Geer.
\newblock {\em Empirical Processes in M-estimation}, volume~6 of {\em Cambridge
  Series in Statistical and Probabilistic Mathematics}.
\newblock Cambridge University Press, 2000.

\bibitem{vapnik_98}
Vladimir Vapnik.
\newblock {\em Statistical learning theory}.
\newblock Wiley, 1998.

\bibitem{villani_03}
C.~Villani.
\newblock {\em Topics in Optimal Transportation}.
\newblock Graduate studies in mathematics. American Mathematical Society, 2003.

\bibitem{vincent_10}
Pascal Vincent.
\newblock A connection between score matching and denoising autoencoders.
\newblock {\em Neural Computation}, 23(7):1661--1674, 2011.

\bibitem{weedbach_17}
J.~Weed and F.~Bach.
\newblock Sharp asymptotic and finite-sample rates of convergence of empirical
  measures in {Wasserstein} distance.
\newblock arXiv preprint 1707.00087, July 2017.

\bibitem{zeileretal_10}
M.~D. Zeiler, D.~Krishnan, G.~W. Taylor, and R.~Fergus.
\newblock Deconvolutional networks.
\newblock In {\em IEEE Computer Society Conference on Computer Vision and
  Pattern Recognition}, pages 2528--2535, June 2010.

\bibitem{zhang_02}
T.~Zhang.
\newblock Covering number bounds of certain regularized linear function
  classes.
\newblock {\em Journal of Machine Learning Research}, 2:527--550, March 2002.

\end{thebibliography}

\appendix
\section{Proofs}\label{sec:proof}
\subsection{Proof of \cref{prop:approx_wass}}
As $\cZ$ is compact, we automatically have $P, f_\sharp \pi \in \cP_2(\Reals^d)$ for any  $\pi \in \cP(\cH)$ and any measurable $f : \Reals^k \to \cZ$. Also, by the measurable selection theorem \cite{rieder_78}, for any $\eps > 0$ there exists a measurable map $\phi_\eps: \cZ \to \cH$ such that $\|z-f(\phi_\eps(z))\|^2 \leq \min_{h \in \cH}\|z - f(h)\|^2 + \eps$ for all $z \in \cZ$. Denote by $\pi_\eps$ the pushforward $(\phi_\eps)_\sharp P$. Evidently, $\pi_\eps \in \cP(\cH)$. Then, since the joint law of $Z$ and $f(\phi_\eps(Z))$ is a coupling of $P$ and $f_\sharp \pi_\eps$, we have
\begin{align*}
	\E_P\min_{h\in\cH}\|Z - f(h)\|^2 + \eps \geq \inf_{M(\cdot\times\cH) = P\atop M(\cZ\times\cdot) = \pi_{\eps}} \E_M\|Z - f(H)\|^2 \geq \inf_{\pi} W_2^2(P,f_\sharp\pi),
\end{align*}
by the definition of $\pi_\eps$. Taking $\eps \to 0$, we get $\risk(P,f) \geq \inf_{Q \in \cF_\sharp\cP(\cH)} W_2^2(P,Q)$. The other direction is straightforward, as for any distribution $\tilde{\pi} \in \cP(\cH)$ and any $z \in \cZ$, we have $\min_{h\in\cH}\|z - f(h)\|^2 \leq \E_{\tilde{\pi}}\|z - f(h)\|^2$.

\subsection{Proof of \cref{prop:rd}}
Consider the following Markov chain:
\begin{align*}
	Z \stackrel{g \in \cG}{\longrightarrow} H \stackrel{f \in \cF}{\longrightarrow} \wh{Z},
\end{align*}
where $Z$ is distributed as $P$, and $\cG$ is a family of all measurable maps $\cZ \to \cH$. Then, we have $I(Z;\wh{Z}) \leq I(Z;H) \leq \log k$, where the first inequality is due to data-processing inequality and the second inequality is by the properties of mutual information \cite{coveretal_12}. Then, we have
\begin{align*}
	\inf_{f \in \cF}\E_P\left[\min_{h \in \cH} \|Z - f(h)\|^2\right] \geq \inf_{f,g: |\cH|\leq k} \E_P\|Z - f(g(Z))\|^2 \geq \underbrace{\inf_{P_{\wh{Z}|Z}: I(Z;\wh{Z}) \leq \log k} \E_P\|Z - \wh{Z}\|^2}_{= D(\log k, P)}.
\end{align*}

\subsection{Proof of \cref{thm:learn_wasserstein}}
Let $\Pi(P_n,P)$ be a set of all couplings of $P_n$ and $P$, i.e., all joint distributions $M \in \cP(\cZ \times \cZ)$, such that $M(\cdot \times \cZ) = P_n$ and $M(\cZ \times \cdot) = P$. Then, for any $M \in \Pi(P_n, P)$ and any admissible decoder $f \in \cF$, we have
\begin{align*}
	\left|\risk(P_n, f) - \risk(P, f)\right| &\leq \int_{\cZ \times \cZ} M(\d z, \d z') \left| \min_{h \in \cH} \|z - f(h)\|^2 - \min_{h' \in \cH}\|z' - f(h')\|^2 \right|\\
	&\leq \int_{\cZ \times \cZ} M(\d z, \d z') \max_{h \in \cH} \left|\|z - f(h)\|^2 - \|z' - f(h)\|^2\right|\\
	&\leq 2\,\diam(\cZ) \int_{\cZ \times \cZ} M(\d z, \d z') \max_{h \in \cH} \left|\|z - f(h)\| - \|z' - f(h)\|\right|\\
	&\leq 2\,\diam(\cZ) \int_{\cZ \times \cZ} M(\d z, \d z') \|z - z'\|,
\end{align*}
where the third inequality uses the identity $\|u\|^2-\|v\|^2 = \langle u+v, u-v\rangle$ and Cauchy--Schwarz. Taking the infimum of both sides over all $M \in \Pi(P_n, P), f \in \cF$, we have
\begin{align*}
	\sup_{f \in \cF}\left|\risk(P_n, f) - \risk(P, f)\right| \leq 2\,\diam(\cZ) \cdot W_1(P_n, P).
\end{align*}
Since both $P$ and $P_n$ are supported on $\cZ$, the value of the function $(Z_1,\ldots,Z_n) \mapsto W_1(P_n,P)$ changes by at most $\frac{1}{n}\diam(\cZ)$ if we replace any $Z_i$ by an arbitrary $z' \in \cZ$. Thus, by McDiarmid's inequality, 
\begin{align}
	\Pr\Big(W_1(P_n,P) - \E W_1(P_n,P) > t \Big) \leq \exp\left(-\frac{2n t^2}{\diam^2(\cZ)}\right). \label{eq:mcdiarmid_w1}
\end{align}
Combining \cref{eq:mcdiarmid_w1} with Wasserstein convergence results of Dereich et al. \cite[Theorems 1--3]{dereichetal_13} (with $p = 1$), we get the claimed result with the constant
\begin{align*}
	C_{q,d} &\deq 18d\cdot2^d \frac{2^{\frac{d-1}{d}q}}{\frac{1}{2}-2^{-\frac{d-1}{d}q}} + 6d\cdot2^{\frac{d}{2}}\frac{2^{\frac{q}{2}}}{1-2^{\frac{2-q}{2}}}.
\end{align*}

\subsection{Proof of \cref{thm:learn_complexity}}
The proof uses a standard chaining argument \cite{vandegeer_00,bartlettetal_17}, except additional care must be taken to relate the properties of the induced class $\cE_\cF \deq \left\{e_f : f \in \cF\right\}$ to those of $\cF$. Given $Z_1, \ldots, Z_n$, define the random process $X_f \deq n^{-1/2}\cdot\sum_{i=1}^n \eps_i \cdot e_f(Z_i)$, where $\{\eps_i\}_{i=1}^n$ are i.i.d.\ Rademacher random variables, i.e., ${\bf P}[\eps_i = \pm 1] = 1/2$, independent of $Z_1,\ldots,Z_n$. By the symmetrization inequality, we have
\begin{align}
	\E_{Z^n} \sup_{f \in \cF} \left[\risk(P,f) - \risk(P_n,f)\right] \leq  \frac{2}{\sqrt{n}} \E_{Z^n}\E_{\eps^n} \sup_{f\in\cF} X_f. \label{eq:symm}
\end{align}
Now, for all $t \in \{0,1,2,\ldots\}$, let $N_t$ be a minimal $(\diam(\cZ)\cdot2^{-t})$-net of $\cF$ in $\|\cdot\|_\cH$ and $\pi_t: \cF \to N_t$ be the corresponding nearest neighbor matching, i.e., $\pi_t(f) \deq \argmin_{f' \in N_t} \|f-f'\|_\cH$. Then, we can telescope $\E\sup_{f\in\cF} X_f$ as
\begin{align}
		\E\sup_{f\in\cF} X_f &\leq \E\sup_{f\in\cF}X_{\pi_0(f)} + \E\sup_{f\in\cF}\left(X_f - X_{\pi_T(f)}\right) + \sum_{t=1}^T\E\sup_{f \in \cF} \left(X_{\pi_t(f)} - X_{\pi_{t-1}(f)} \right), \label{eq:telescoping}
\end{align}
for some $T \in \mathbb{N}$ (to be tuned later). Since $|N_0| = 1$ (we can take any singleton $\{f\} \subset \cF$ to be a minimal $\diam(\cZ)$-net of $\cF$), the first term is zero. To handle the remaining two terms, we will need the following estimate: for any $z \in \cZ$ and $f,f' \in \cF$,
\begin{align}\label{eq:ef-ef'}
	|e_f(z)-e_{f'}(z)| \le 2\,\diam(\cZ) \cdot \|f-f'\|_\cH.
\end{align}
To prove this inequality, we write
\begin{align*}
	|e_f(z) - e_{f'}(z)| &= \left|\min_{h \in \cH}\max_{h' \in \cH} \left(\|z-f(h)\|^2-\|z-f'(h')\|^2\right)\right| \\
	&\le \max_{h \in \cH} \left|\|z-f(h)\|^2 - \|z-f'(h)\|^2\right| \\
	&= \max_{h \in \cH} \left|\langle (z-f(h)) + (z-f'(h)), f'(h)-f(h) \rangle\right| \\
	&\le 2\,\diam(\cZ) \cdot \| f - f' \|_\cH.
\end{align*}
Now we can estimate the second term in \eqref{eq:telescoping} as follows:
\begin{align*}
	\E_{\eps^n} \sup_{f\in\cF} (X_f - X_{\pi_T(f)}) &= \frac{1}{\sqrt{n}}\E_{\eps^n}\sup_{f\in\cF} \sum_{i=1}^n \eps_i \left(e_f(Z_i) - e_{\pi_T(f)}(Z_i)\right)\\
	&\leq \frac{1}{\sqrt{n}}\E_{\eps^n}\sqrt{\sum_{i=1}^n \eps_i^2}\sqrt{\sup_{f\in\cF}\sum_{i=1}^n \left(e_f(Z_i) - e_{\pi_T(f)}(Z_i)\right)^2}\\
	&\leq 2\sqrt{n}\cdot\diam^2(\cZ)2^{-T},
\end{align*}
where the first inequality is by Cauchy--Schwarz, while the second inequality follows from \eqref{eq:ef-ef'} applied to $f' = \pi_T(f)$. For the third term of \cref{eq:telescoping}, we have for any $t \in {\mathbb{N}}$
\begin{align*}
		\left|e_{\pi_t(f)} - e_{\pi_{t-1}(f)}\right| &\leq 2\,\diam(\cZ)\cdot \|\pi_t(f) - \pi_{t-1}(f)\|_\cH\\
		&\leq 2\,\diam(\cZ)\cdot \left(\|\pi_t(f) - f\|_\cH + \|f - \pi_{t-1}(f)\|_\cH\right)\\
		&\leq 6\,\diam^2(\cZ)\cdot 2^{-t}.
\end{align*}
By Hoeffding's lemma, it follows that each $X_{\pi_t(f)} - X_{\pi_{t-1}(f)}$ is a $36\,\diam^4(\cZ)2^{-2t}$-subgaussian random variable. By using the maximal inequality for subgaussian random variables \cite[Sec.~2.5]{boucheronetal_13}, we can upper bound \cref{eq:telescoping} by
\begin{align*}
		\E\sup_{f\in\cF} X_f &\leq 2\sqrt{n}\cdot \diam^2(\cZ)2^{-T} + 12\,\diam^2(\cZ)\sum_{t=1}^T 2^{-t} \sqrt{\log \cN(\cF,\|\cdot\|_\cH,\diam(\cZ) 2^{-t})}.
\end{align*}
Turning into the entropy integral form:
\begin{align*}
		\E\sup_{f\in\cF} X_f &\leq 2\sqrt{n}\cdot \diam^2(\cZ)2^{-T} + 24\, \diam(\cZ) \int_{\diam(\cZ)\cdot2^{-T-1}}^{\diam(\cZ)/2} \sqrt{\log\cN(\cF,\|\cdot\|_{\cH},u)} d u.
\end{align*}
Selecting $T = \lceil \log_2\left(\diam(\cZ)/2\alpha\right)\rceil$, and plugging into \cref{eq:symm}, we get
\begin{align*}
	\E_{Z^n} \sup_{f \in \cF} \left[\risk(P,f) - \risk(P_n,f)\right] \leq \frac{48}{\sqrt{n}}\diam(\cZ)\cdot\comp(\cF).
\end{align*}
By combining with McDiarmid's inequality (see \cref{eq:mcdiarmid_w1}), and handling the other direction $\sup_{f\in\cF} \risk(P_n, f) - \risk(P,f)$ similarly, we get what we want.

\subsection{Proof of \cref{thm:dense}}
Let $f_\ell(\cdot) = f(\cdot;A_{1:\ell}) \in \cF_\ell$ be a neural net with weight matrices $A_{1:\ell}$. For each $j \in \{1,\ldots,\ell\}$, we will use the shorthand notation $F_j$ for the layerwise transformation $F_j(\cdot;A_j)$, so that $\pi_\cZ \circ f_\ell = \pi_\cZ \circ F_\ell \circ F_{\ell-1} \circ \ldots \circ F_1(h)$. Then, using the fact that $\cZ \ni {\mathbf 0}$, we can write
\begin{align}
	\|\pi_\cZ (f_\ell(h)) \| &= \|\pi_\cZ(f_\ell(h)) - \pi_\cZ(f_\ell(\mathbf{0})) \| \nonumber\\
	&\le \| f_\ell(h) - f_\ell(\mathbf{0}) \| \nonumber\\
	&= \|\sigma_\ell(A_\ell F_{\ell-1}(h)) - \sigma_\ell(A_\ell\mathbf{0})\| \nonumber\\
	&\leq L_\ell \|A_\ell\|\|F_{\ell-1}(h)\| \nonumber \\
	&\leq \left(\prod_{j=1}^\ell L_j M_j\right) \|h\|, \label{eq:recursion_1}
\end{align}
where we have used the fact that the projection map $\pi_\cZ$ onto a closed convex set $\cZ$ is nonexpansive, i.e., $\|\pi_\cZ(u)-\pi_\cZ(v)\| \le \|u-v\|$ for all $u,v$, and where
the last inequality follows from the relationship $\|A\| \leq \|A\|_2 \leq \|A\|_1$ and from applying the same argument recursively. Also note that $\|\pi_\cZ(f_\ell(h))\| \leq \diam(\cZ)$, since $\pi_\cZ$ projects $f_\ell(h)$ onto $\cZ$. Then it follows that the diameter of the class $\cF_\ell$ in $\|\cdot\|_\cH$, i.e., $\sup_{f,f' \in \cF_\ell}\|f - f'\|_\cH$, is bounded from above by $2(\prod_{j=1}^\ell L_j M_j)\|\cH\|_\infty =: 2D$. Now let $\tilde{f}_\ell$ be another neural net with matrices $\tilde{A}_{1:\ell}$, such that $\pi_\cZ \circ \tilde{f}_\ell = \pi_\cZ \circ \tilde{F}_\ell \circ \tilde{F}_{\ell-1} \circ \ldots \circ \tilde{F}_1$. Then, using the nonexpansiveness of the projection $\pi_\cZ$ and Lipschitz continuity again, we can proceed as
\begin{align}
	\|\pi_\cZ(f_\ell(h))-\pi_\cZ(\tilde{f}_\ell(h))\| &\leq \|\sigma_\ell(A_\ell F_{\ell-1}(h)) - \sigma_\ell(\tilde{A}_\ell \tilde{F}_{\ell-1}(h))\| \nonumber\\
	&\leq L_\ell  \|A_\ell F_{\ell-1}(h) - \tilde{A}_\ell F_{\ell-1}(h)\| + L_\ell \| \tilde{A}_\ell F_{\ell-1}(h) - \tilde{A}_\ell \tilde{F}_{\ell-1}(h)\| \nonumber\\
	&\leq L_\ell \|A_\ell - \tilde{A}_\ell\| \cdot \|F_{\ell-1}(h)\| + L_\ell M_\ell \|F_{\ell-1}(h) - \tilde{F}_{\ell-1}(h)\| \nonumber\\
	&\leq D \sum_{j=1}^\ell \frac{\|A_j - \tilde{A}_j\|}{M_j}, \label{eq:recursion_2}
\end{align}
where the last inequality follows by \cref{eq:recursion_1} and recursion. From \cref{eq:recursion_2}, we see that the covering number of $\cF$ in $\|\cdot\|_\cH$ can be estimated as
\begin{align*}
	\cN(\cF,\|\cdot\|_\cH,\eps) \leq \prod_{j=1}^\ell \cN\left(\frac{\cA_j}{M_j},\|\cdot\|,\frac{\omega_j \eps}{D}\right) \leq \prod_{j=1}^\ell \cN\left(\frac{\cA_j}{M_j},\|\cdot\|_2,\frac{\omega_j \eps}{D}\right),
\end{align*}
for any choice of positive weights $\omega_1,\ldots,\omega_\ell$ summing up to $1$, where $\cA_j/M_j \deq \left\{ A_j/M_j : A_j \in\cA_j \right\}$. Then, for any $\alpha > 0$ and weights $\omega_1,\ldots,\omega_\ell > 0$, the entropy integral $\comp(\cF)$ can be upper bounded as follows:
\begin{align*}
	\comp(\cF) &\leq \frac{\alpha\sqrt{n}}{6} + \int_{\alpha}^{\frac{\diam(\cZ)}{2}} \sqrt{\sum_{j=1}^\ell \log \cN\left(\frac{\cA_j}{M_j},\|\cdot\|_2,\frac{\omega_j \eps}{D}\right)} \d \eps,
\end{align*}
Selecting the weights $\omega_j = 1/\ell$ and simplifying further,
\begin{align*}
	\comp(\cF) &\leq \frac{\alpha\sqrt{n}}{6} + \int_{\alpha}^{\frac{\diam(\cZ)}{2}} \sqrt{\sum_{j=1}^\ell \log \cN\left(\frac{\cA_j}{M_j},\|\cdot\|_2,\frac{\eps}{\ell D}\right)} \d \eps\\
	&\leq \frac{\alpha\sqrt{n}}{6} + \int_{\alpha}^{\frac{\diam(\cZ)}{2} \wedge \ell\cdot D} \sqrt{\left\lceil \frac{\ell^2 D^2}{\eps}\right\rceil} \sqrt{\sum_{j=1}^\ell\log \left(1+\frac{2w_{j-1}w_0 \eps^2}{\ell^2D^2}\right)} \d \eps\\
	&\leq \frac{\alpha\sqrt{n}}{6} + (\ell D\sqrt{2})\sqrt{\sum_{j=1}^\ell\log \left(1+2w_{j-1}w_0\right)} \cdot \int_{\frac{\alpha}{\ell D}}^{\frac{\diam(\cZ)}{2\ell D} \wedge 1} \frac{1}{u}  \d u,
\end{align*}
where for the second inequality we used the Maurey-type bounds on the covering numbers (see, e.g., \cite{zhang_02} or \cref{ssec:sparsification} for a short derivation) and for the last inequality we used the substitution $u = \ell D\cdot \eps$ and the fact that $\lceil x \rceil \leq 2x$ for $x \geq 1$. Evaluating the integral with the choice $\alpha = \diam(\cZ)/2\sqrt{n}$, we get the claimed bound.

\subsection{Proof of \cref{thm:dense_deep}}\label{ssec:pf_dense_deep}
First, note that for any $f = \pi_\cZ \circ f_\ell \in \cF_\ell$, $e_{f}(z) \in [0,\diam^2(\cZ)]$ for any $z \in \cZ$, as $\|z - \pi_\cZ (f_\ell(h))\|^2 \leq \diam^2(\cZ)$ holds for any $z \in \cZ$ and $h \in \cH$. Moreover, the estimate \cref{eq:recursion_2} from the proof of \cref{thm:dense} still holds (again, let $D \deq \|\cH\|_\infty (\prod_{j=1}^\ell L_j M_j)$). Now, we use the volumetric covering number estimates for balls in finite-dimensional Banach spaces \cite{chafai_12} to proceed as
\begin{align*}
	\cN(\cF_\ell,\|\cdot\|_{\cH},\eps) \leq \prod_{j=1}^\ell \cN\left(\frac{\cA_j}{M_j},\|\cdot\|,\frac{\omega_j \eps}{D}\right) \leq \prod_{j=1}^\ell \left(\frac{3D}{\omega_j \eps}\right)^{w_{j-1}w_j}.
\end{align*}
With the (suboptimal) choice $\omega_j = 1/\ell$, we get $\cN(\cF_\ell,\|\cdot\|_\cH,\eps) \leq (3\ell D/\eps)^{\sum_{j=1}^\ell w_{j-1}w_j}$. Combining this with Lemma 21 of \cite{vainsencheretal_11} (see \cref{ssec:lemma21}), we get the claimed result.

\subsection{Proof of \cref{thm:cnn_1d}}\label{ssec:pf_cnn_1d}
First, notice that the convolution operation is linear in both $\xi$ and $A$, so that, for pairs of inputs $\xi, \tilde{\xi}$ and convolution filters $A^{(1:v)}, \tilde{A}^{(1:v)}$, we have
\begin{align}
	\|\conv(\xi;A^{(1:v)}) - \conv(\tilde{\xi};\tilde{A}^{(1:v)})\|_2 &\leq \|\conv(\xi;A^{(1:v)} - \tilde{A}^{(1:v)})\|_2 + \|\conv(\xi - \tilde{\xi};\tilde{A}^{(1:v)})\|_2. \label{eq:triangle_cnn}
\end{align}
Now, we show that the convolution inequality \cref{eq:young_ineq} holds: for an input $\xi \in \Reals^{w_0 \times v_0}$ a mapping $\conv(\cdot; A^{(1:v)}) : \Reals^{w_0 \times v_0} \to \Reals^{w \times v}$ with $v$ channels of convolution filters $A^{(k)} \in \Reals^{u \times v_0}$,
\begin{align*}
	\|\conv(\xi;A^{(1:v)})\|_2^2 &\leq \sum_{j=1}^v \sum_{i=1}^w\left(\sum_{i'=1}^u\sum_{j'=1}^{v_0} \sqrt{|A^{(j)}_{i',j'}|}\sqrt{|A^{(j)}_{i',j'}|}\left|\xi_{i'+s(i-1)+\frac{1-u}{2},j'}\right|\right)^2\\
	&\leq \sum_{j=1}^v \sum_{i=1}^w \left\|A^{(j)}\right\|_1 \cdot \left[\sum_{i'=1}^u \sum_{j'=1}^{v_0}\left|A^{(j)}_{i',j'}\right|\left|\xi_{i'+s(i-1)+\frac{1-u}{2},j'}\right|^2\right]\\
	&\leq \left(\sum_{j=1}^v \left\|A^{(j)}\right\|^2_1\right) \cdot \max_{i' \in [u], j' \in [v_0]}\left(\sum_{i=1}^w \left|\xi_{i'+s(i-1)+\frac{1-u}{2},j'}\right|^2\right)\\
	&\leq \left(\sum_{j=1}^v \left\|A^{(j)}\right\|_1^2\right) \cdot \|\xi\|_2^2,
\end{align*}
where we have used the Cauchy--Schwarz inequality in the second step and H\"older's inequality in the third step. Taking the square root of each side, we get \cref{eq:young_ineq}.

Now, analogously to \cref{eq:recursion_2}, we can proceed by combining \cref{eq:triangle_cnn} and \cref{eq:young_ineq}. First, define the norm $\|A^{(1:v)}\|_{1,2} \deq \sqrt{\sum_{k=1}^v \|A^{(k)}\|_1^2}$ for $v$ channels of convolution matrices. Then, for any $f_\ell, \tilde{f}_\ell$ indexed by the filter weights $\{A^{(1:v_j)}_{j}\}_{j=1}^\ell, \{\tilde{A}^{(1:v_j)}_j\}_{j=1}^\ell$, we have
\begin{align*}
	\left\|\pi_\cZ(f_\ell(h)) - \pi_\cZ(\tilde{f}_\ell(h))\right\| &\leq \left\|A_{\ell}^{(1:v_\ell)} - \tilde{A}_\ell^{(1:v_\ell)}\right\|_{1,2} \|F_{\ell-1}(h)\|_2 + M_\ell\left\|F_{\ell-1}(h) - \tilde{F}_{\ell-1}(h)\right\|_2\\
	&\leq D\sum_{j=1}^\ell \frac{\left\|A_{j}^{(1:v_j)} - \tilde{A}_j^{(1:v_j)}\right\|_{1,2}}{M_j}
\end{align*}
where $D = \|\cH\|_\infty \cdot \prod_{j=1}^\ell M_j$. The remaining steps are identical to those in the proof of \cref{thm:dense_deep} (see \cref{ssec:pf_dense_deep}), by invoking the bound on the covering numbers in the normed spaces $(\Reals^{u_jv_{j-1}v_j},\|\cdot\|_{1,2})$.

\subsection{Proof of \cref{thm:cnn_ld}}
The proof is same as the proof of \cref{thm:cnn_ld}, except that we need a higher-order version of Young's convolution inequality. For an input $\xi \in \Reals^{w_{0,1} \times \cdots \times w_{0,m} \times v_0}$, $v$ channels of convolution weight tensors $A^{(k)} \in \Reals^{u_{1} \times \cdots \times u_{m} \times v_{0}}$, and filter strides $s_1,\ldots,s_m$, we have
\begin{align*}
	&\|\conv(\xi;A^{(1:v)})\|_2^2\\
	&\leq \sum_{j=1}^v \sum_{i_{1:m}=1}^{w_{1:m}} \left(\sum_{i'_{1:m}=1}^{u_{1:m}}\sum_{j'=1}^{v_0} \left(\sqrt{|A^{(j)}_{i'_1,\ldots,i'_m,j'}|}\right)^2\left|\xi_{i'_1+s_1(i_1-1)+\frac{1-u_1}{2},\ldots,i'_m+s_m(i_m-1)+\frac{1-u_m}{2},j'}\right|\right)^2\\
	&\leq \sum_{j=1}^v \sum_{i_{1:m}=1}^{w_{1:m}}\|A^{(j)}\|_1 \cdot\left[\sum_{i'_{1:m}=1}^{u_{1:m}} \sum_{j'=1}^{v_0}\left|A^{(j)}_{i'_1,\ldots,i'_m,j'}\right|\left|\xi_{i'_1+s_1(i_1-1)+\frac{1-u_1}{2},\ldots,i'_m+s_m(i_m-1)+\frac{1-u_m}{2},j'}\right|^2 \right]\\
	&\leq \left(\sum_{j=1}^v \|A^{(j)}\|_1^2\right)\cdot\max_{i'_{1:m} \in [u_{1:m}]\atop j' \in [v_0]}\left(\sum_{i_{1:m}=1}^{w_{1:m}} |\xi_{i'_1+s_1(i_1-1)+\frac{1-u_1}{2},\ldots,i'_m+s_m(i_m-1)+\frac{1-u_m}{2},j'}|^2\right)\\
	&\leq \left(\sum_{j=1}^v \|A^{(j)}\|_1^2\right)\cdot \|\xi\|_2^2,
\end{align*}
analogously to the procedure in \cref{ssec:pf_cnn_1d}, where we have introduced the following shorthand notations: we use $\sum_{i_{1:m}=1}^{w_{1:m}}$ to denote $\sum_{i_1=1}^{w_1}\cdots \sum_{i_m=1}^{w_m}$, and use $\max_{i'_{1:m} \in [u_{1:m}]}$ to denote $\max_{i'_1 \in [u_1]} \cdots \max_{i'_m \in [u_m]}$. Then we can proceed as in \cref{ssec:pf_dense_deep}.

\subsection{Proof of \cref{thm:dense_sigmoid}}
Similar to \cref{eq:recursion_2}, we proceed as follows: for any $f_\ell, \tilde{f}_\ell$ indexed by $A_{1:\ell}$, $\tilde{A}_{1:\ell}$, we have
\begin{align*}
	\|\pi_\cZ(f_\ell(h)) - \pi_\cZ(\tilde{f}_\ell(h))\| &\leq L_\ell\cdot\|A_\ell - \tilde{A}_\ell\|\cdot\|F_{\ell-1}(h)\| + L_\ell \|\tilde{A}_\ell\| \|F_{\ell-1}(h) - \tilde{F}_{\ell-1}(h)\|\\
	&\leq L_\ell B_{\ell-1}\|A_\ell - \tilde{A}_\ell\| + L_\ell M_\ell \|F_{\ell-1}(h) - \tilde{F}_{\ell-1}(h)\|\\
	&\leq \sum_{i=1}^\ell \left(\prod_{j=i}^\ell L_j M_j\right) B_{i-1}\frac{\|A_i - \tilde{A}_i\|}{M_i},
\end{align*}
where the last inequality is by recursion, with $B_0 \deq \|\cH\|_\infty$. We now use the shorthand notation $D_i \deq B_{i-1}(\prod_{j=i}^\ell L_j M_j)$. For any choice of weights $\omega_1, \ldots, \omega_\ell > 0$, we can upper-bound the covering number as
\begin{align*}
	\cN(\cF_\ell,\|\cdot\|_\cH,\eps) \leq \prod_{i=1}^\ell \cN\left(\frac{\cA_i}{M_i},\|\cdot\|,\frac{\omega_i \cdot \eps}{D_i}\right) &\leq \prod_{i=1}^\ell \cN\left(\frac{\cA_i}{M_i},\|\cdot\|_2,\frac{\omega_i \cdot \eps}{D_i}\right),
\end{align*}
using the relationship of the operator norm and the entrywise $\ell_2$-norm. Now, we choose $\omega_i = D_i/\sum_{j=1}^\ell D_j$ and invoke Maurey's empirical method (\cref{ssec:sparsification}) to proceed as
\begin{align*}
	\log \cN(\cF_\ell,\|\cdot\|_\cH,\eps) &\leq \left\lceil\frac{(\sum_{j=1}^\ell D_j)^2}{\eps^2}\right\rceil \cdot \sum_{i=1}^\ell  \log\left(1+\frac{2w_{i-1}w_i \eps^2}{(\sum_{j=1}^\ell D_j)^2}\right),
\end{align*}
for $\eps \leq \sum_{j=1}^\ell D_j$ (otherwise, the covering number is $1$). Evaluating the entropy integral with the choice $\alpha = \diam(\cZ)/2\sqrt{n}$ and plugging the estimate into the \cref{thm:learn_complexity}, we get the claimed bound.

\subsection{Covering number bounds based on Maurey's empirical method} \label{ssec:sparsification}
Here, we provide a short derivation of an upper bound \cref{eq:jbm} on the covering number of an $\ell_1$ ball by smaller $\ell_2$ balls with radius $\eps$. The proof goes through the standard sparsification steps (see \cite{zhang_02} and references therein), and is included only for completeness.

First note that we can assume that the radius of the $\ell_1$ ball (denoted henceforth as $\cB_1$) to be $1$ without loss of generality, as we can rescale the $\ell_2$ balls to have radius $\eps/M$. Let $\{e_1,\ldots,e_d\}$ be the standard basis of $\Reals^d$. Now, for an arbitrary $v \in \cB_1$, let $U$ be a random vector in $\Reals^d$ constructed as
\begin{align*}
	U = \begin{cases} \mathsf{sign}(v_i) e_i, & \text{w.p. } |v_i|, \forall i \in \{1,\ldots,d\}\\ 0, & \text{w.p. } 1-\|v\|_1 \end{cases},
\end{align*}
satisfying $\E U = v$. Let $U_{(1)}, \ldots, U_{(k)}$ be i.i.d.\ copies of $U$ for some fixed $v$ and some $k \in \mathbb{N}$ (to be tuned later), and let $\bar{U} = \frac{1}{k}\sum_{j=1}^k U_{(j)}$. Then,
\begin{align*}
	\E\|\bar{U} - v\|^2 = \sum_{i=1}^d \E\|\bar{U_i} - v_i\|^2 = \frac{1}{k}\sum_{i=1}^d \E(U_i - v_i)^2 \leq \frac{1}{k},
\end{align*}
where the last inequality holds as $\E(U_i - v_i)^2 = |1-v_i|\cdot|v_i| \leq |v_i|$. If we choose $k = \lceil 1/\eps^2\rceil$, then $\E\|\bar{U} - v\|^2 \leq \eps^2$, which implies that there is at least one realization of $\bar{U}$, such that $\|\bar{U}-v\| \leq \eps$. As the number of distinct values that $\bar{U}$ can take is upper bounded by $(2d+1)^k$ (irrespective of the choice of $v$), we get what we want.

\subsection{A high-probability uniform deviation bound}\label{ssec:lemma21}

For completeness, we state Lemma 21 of Vainsencher et al. \cite{vainsencheretal_11} based on the single covering step, which has been referred to in the discussion following \cref{thm:learn_complexity} and the proof in \cref{ssec:pf_dense_deep}. Note that the lemma has been slightly adapted for the sake of notational coherence.
\begin{lemma}
Let $\cG$ be a class of functions $g:\cZ \to [0,B]$ with the covering number bound
\begin{align*}
	\cN(\cG,\|\cdot\|_{\infty},\eps) \leq \left(\frac{C}{\eps}\right)^d
\end{align*}
for some constant $d, C$, whenever $(C/\eps)^d > e/B^2$ holds. Then, for every $\delta \in (0,1)$,
\begin{align*}
	\sup_{g\in\cG}\left[\E_{P} g(Z) - \E_{P_n} g(Z)\right] \leq B\left(\sqrt{\frac{d \ln (C\sqrt{n})}{2n}} + \sqrt{\frac{\log(1/\delta)}{2n}}\right) + \frac{2}{\sqrt{n}}.
\end{align*}
\end{lemma}

\end{document}